\documentclass{article}

\PassOptionsToPackage{numbers, compress}{natbib}
\usepackage[main,final]{neurips_2026}

\usepackage[utf8]{inputenc} % allow utf-8 input
\usepackage[T1]{fontenc}    % use 8-bit T1 fonts
\usepackage{hyperref}       % hyperlinks
\usepackage{url}            % simple URL typesetting
\usepackage{booktabs}       % professional-quality tables
\usepackage{amsfonts}       % blackboard math symbols
\usepackage{nicefrac}       % compact symbols for 1/2, etc.
\usepackage{microtype}      % microtypography
\usepackage[table]{xcolor}       % colors
\usepackage{amsmath}
\usepackage{graphicx}
\usepackage{makecell}
\usepackage{multirow}
\usepackage[most]{tcolorbox}

\newtcolorbox{promptbox}[1]{%
    colback=gray!5,        % 背景颜色
    colframe=gray!75,      % 边框颜色
    fonttitle=\bfseries,   % 标题字体
    title=#1,              % 标题内容
    arc=2mm,               % 圆角
    boxrule=0.5pt,         % 边框粗细
    left=3mm, right=3mm, top=2mm, bottom=2mm, % 内边距
    enhanced,
    breakable              % 支持跨页
}

\newcommand{\method}{VicEdit}
\newcommand{\dataset}{VicEdit-400K}
\newcommand{\benchmark}{VicEdit-Bench}

\title{\method{}: Learning to Edit Videos from Visual In-Context Examples}

\author{%
  Yuji Wang$^{1*}$ ~
  Teng Hu$^{1}$\thanks{Equal contributions.} ~
  Yuheng Chen$^{1}$ ~
  Ran Yi$^{1}$\thanks{Project lead.} ~
  Han Feng$^{2}$ ~
  Weijian Cao$^{2}$ \\
  \textbf{Chengjie Wang}$^{2}$ ~
  \textbf{Lizhuang Ma}$^{1}$\thanks{Corresponding author.} ~
  \textbf{Jiangning Zhang}$^{3}$ ~ \\
  \normalsize $^1$Shanghai Jiao Tong University ~~ $^2$Tencent Youtu Lab ~~ $^3$Zhejiang University \\
\noindent \texttt{Project Page:} \url{https://rain152.github.io/VicEdit/} \\
}

\begin{document}
\maketitle

\begin{abstract}
Despite remarkable progress in instruction-based video editing, textual descriptions remain limited in expressing fine-grained appearance, spatial transformations, and temporal dynamics. We introduce Visual In-context Editing, a new paradigm that shifts video editing from text-only instructions to visual demonstrations, where users specify desired edits through heterogeneous visual references including single images, image pairs, and video pairs. To establish this paradigm, we curate \dataset{}, the first large-scale dataset for visual in-context video editing, containing 400K high-quality samples across ten editing tasks with comprehensive semantic and visual consistency filtering. Building upon this foundation, we propose \method{}, a unified framework that enables effective reasoning over diverse visual contexts and textual instructions. Specifically, we introduce Modality-Adaptive Semantic Distillation to extract modality-aware visual editing semantics and Dual-Context Injection to align visual contexts with textual editing intents. Extensive experiments on \benchmark{} demonstrate that \method{} achieves state-of-the-art performance on both conventional instruction-based editing and visual in-context editing. Further controlled studies and human evaluations verify that \method{} effectively leverages heterogeneous visual references, establishing visual in-context editing as a promising paradigm for controllable video editing.

\end{abstract}
\section{Introduction}
\label{sec:introduction}

\begin{figure}[t]
\centering
\includegraphics[width=0.9\linewidth]{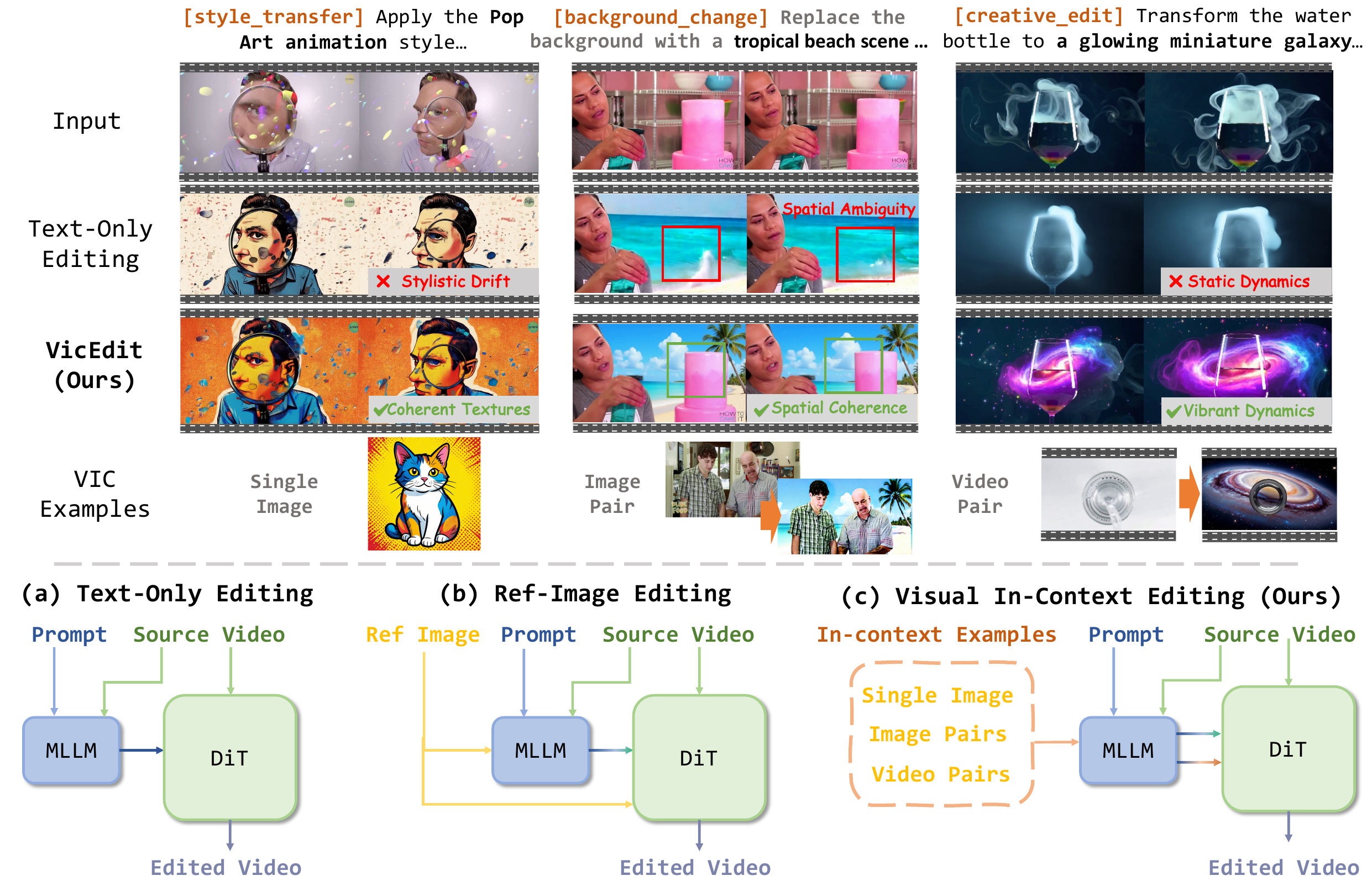}
\caption{
\textbf{Top:} Qualitative comparison of \method{} with text-only baselines. Single-image, image-pair, and video-pair references help mitigate stylistic drift, spatial inaccuracies, and insufficient motion guidance, respectively.
\textbf{Bottom:} Video editing paradigms: (a) text-only guidance; (b) visual reference conditioning via auxiliary branches; and (c) \method{}, which distills modality-aware editing semantics from heterogeneous references and fuses them with text instructions within the video DiT.
All examples are from OpenVE~\cite{he2025openve}.
}
\vspace{-2mm}
\label{fig:teaser}
\end{figure}

Instruction-based video editing~\cite{mai2025easyv2v, pan2026nova, yang2026videocof, kim2025temporal, sun2024survey} has made remarkable progress, enabling users to modify video content through natural language with increasing controllability. Built upon advances in large language models and diffusion architectures~\cite{zheng2024open,wan2025wan,kong2024hunyuanvideo}, recent methods support a broad range of editing operations, from localized object manipulation to global style transfer. However, textual instructions remain limited in expressing the fine-grained visual details, precise spatial transformations, and complex temporal dynamics required for high-fidelity editing. Describing a desired appearance, an object's exact transformation, or a complex motion pattern in words can be challenging, even when the intended result is visually clear.

As illustrated in Fig.~\ref{fig:teaser} (Top), heterogeneous visual references offer complementary forms of editing guidance. A \textit{Single Image} provides direct cues about the desired appearance and style, an \textit{Image Pair} illustrates the visual and spatial changes between two states, and a \textit{Video Pair} conveys temporal patterns that are difficult to specify precisely through text alone. Although existing methods have explored text-only and reference-guided editing, their ability to exploit heterogeneous visual references remains limited. Moreover, existing benchmarks~\cite{lin2026kiwi, zhang2025reco, he2025openve} provide insufficient coverage of diverse reference modalities for systematically studying this setting. These limitations motivate a shift from describing edits through text alone to demonstrating them through visual examples, which we formulate as \textbf{Visual In-context Editing}.

Visual in-context editing is a unified video editing paradigm in which visual demonstrations complement textual instructions to communicate the desired edit. Rather than restricting users to a single reference modality, it accommodates Single Image, Image Pair, and Video Pair inputs, enabling different forms of visual evidence to guide the editing process. Realizing this paradigm presents two key challenges: extracting relevant editing semantics from heterogeneous references, whose information varies across modalities, and integrating these visual semantics with textual instructions while preserving their complementary roles.

To address these challenges, we introduce \textbf{\dataset{}} (Sec.~\ref{sec:data}), the first large-scale dataset specifically designed for visual in-context video editing. It comprises 400K high-quality samples across three reference modalities and ten editing tasks, curated through an automated pipeline with structured instruction generation and quality filtering. Building upon this dataset, we propose \textbf{\method{}} (Sec.~\ref{sec:methodology}), a unified framework for editing with heterogeneous visual references. Specifically, \textbf{Modality-Adaptive Semantic Distillation (MASD)} uses modality-conditioned learnable queries to extract modality-aware editing semantics from visual references. \textbf{Dual-Context Injection (DCI)} then integrates the resulting visual context tokens with textual instructions into the video diffusion transformer (DiT), enabling the generation process to leverage both sources of editing information.

We evaluate \method{} on \textbf{\benchmark{}}, which assesses visual in-context editing across the three reference modalities, as well as on conventional instruction-based video editing tasks. Extensive experiments demonstrate that \method{} achieves state-of-the-art performance in both settings. Controlled studies and human evaluations further examine its ability to leverage heterogeneous visual references, while ablation studies validate the contributions of its key components.

Our contributions are summarized as follows:

\hspace*{1em} \textbf{\textit{1)}} We formulate a unified visual in-context video editing paradigm, transitioning from descriptions to visual demonstrations. To support this, we introduce \textbf{\dataset{}}, the first large-scale dataset featuring 400K high-quality samples across three reference modalities and ten tasks, providing a systematic foundation for visual in-context editing.

\hspace*{1em} \textbf{\textit{2)}} We propose the \textbf{\method{}} framework designed for heterogeneous visual references. Specifically, we introduce \textbf{Modality-Adaptive Semantic Distillation} to adaptively extract editing semantics via modality-conditioned queries, and \textbf{Dual-Context Injection} to synergistically fuse distilled visual tokens with textual instructions into DiT via cross-attention.

\hspace*{1em} \textbf{\textit{3)}} Experiments demonstrate that \method{} not only excels in base text-only editing but also shows significant superiority on our proposed \textbf{\benchmark{}}, which comprehensively evaluates visual in-context editing across three reference modalities. Extensive ablation studies further validate the efficacy of each architectural component.

\section{Related Work}
\textbf{Instruction-driven Video Editing.}
Instruction-driven video editing has evolved from training-free propagation~\cite{wu2023tuneavideo, geyer2023tokenflow, ku2024anyv2v} to large-scale trained systems~\cite{mai2025easyv2v, pan2026nova, yang2026videocof, zhang2025reco}. Recently, several works have begun exploring visual references: Kiwi-Edit~\cite{lin2026kiwi} introduces single reference images via fixed queries, OmniTransfer~\cite{zhang2026omnitransfer} utilizes video references for generation, and EditVerse~\cite{ju2025editverse} encodes task types through in-context tokens. Despite these efforts, a unified framework supporting heterogeneous reference modalities remains absent. \method{} addresses this gap by proposing a visual in-context paradigm that extends editing from text-only instructions to multi-modal guidance across single images, image pairs, and video pairs. To support this paradigm, we construct \dataset{}, the first large-scale dataset covering three reference modalities and ten task types with 400K samples.

\textbf{In-Context Learning in Video Generation.}
In-context learning (ICL)~\cite{dong2024icl-survey} has emerged as a powerful paradigm where models learn to perform tasks by conditioning on demonstrations rather than explicit specifications~\cite{dong2024icl-survey}. In image generation, methods like Painter~\cite{wang2023images} and ImageBrush~\cite{yang2023imagebrush} demonstrate that visual exemplars enable more precise control than text alone. This paradigm has since extended to video: early works~\cite{fei2024video} explored prompt-level ICL for multi-scene tasks, while VICL~\cite{zhang2025vicl} utilized an autoregressive DiT for viewpoint and motion transfer. Recent methods like EditVerse~\cite{ju2025editverse,kim2025temporal} leverage global attention for context interaction, yet struggle with long-sequence references. Alternatively, OmniTransfer~\cite{zhang2026omnitransfer} employs an auxiliary DiT to learn from reference examples. In summary, existing methods are typically confined to video generation and support only limited reference modalities. \method{} advances this field by introducing Modality-Adaptive Semantic Distillation to adaptively extract editing semantics from diverse references, and Dual-Context Injection to synergistically integrate these visual cues with textual instructions within the DiT latent space.

\section{Method}
\label{sec:method}
\subsection{Formulation: Visual In-Context Editing}
\label{sec:formulation}
We define the task of visual in-context video editing as a conditional generation process~\cite{ju2025editverse,fei2024video}. Given a source video $V_{src} \in \mathbb{R}^{T \times 3 \times H \times W}$ and a textual instruction $P$, the goal is to synthesize an edited video $V_{tgt}$ that manifests the editing intentions through an optional visual reference $R$. Formally, this transformation is represented as:
\begin{equation}
V_{tgt} = \mathcal{G}(V_{src}, P, \langle R \rangle),
\end{equation}
where $\mathcal{G}$ denotes the generative model and $\langle \cdot \rangle$ represents an optional conditioning input~\cite{fei2024video, tan2025omnivideo}. To encompass diverse editing scenarios, $R$ can take one of three reference modalities, each conveying complementary editing semantics:
\begin{itemize}
    \item \textbf{Single Image} ($R_{si} = I_{ref}$): Provides fine-grained visual anchors, including textures and appearances, to guide stylistic or identity-consistent synthesis. 
    \item \textbf{Image Pair} ($R_{ip} = \{I_{s}^{ref}, I_{t}^{ref}\}$): Characterizes the transformation between reference domains, capturing shifts in spatial layouts and localized details.
    \item \textbf{Video Pair} ($R_{vp} = \{V_{s}^{ref}, V_{t}^{ref}\}$): Provides dense temporal priors and dynamic semantic correspondences for motion-intensive editing.
\end{itemize}

This paradigm faces two challenges: the scarcity of large-scale unified datasets across these modalities and the structural heterogeneity that complicates semantic extraction and multimodal integration. 

\subsection{Data Curation for Visual In-Context Editing}
\label{sec:data}
In this section, we present the construction pipeline of \dataset{}. We curate 400K high-quality samples by synthesizing editing instructions, generating corresponding visual references, and applying multi-stage quality filtering. Based on this dataset, we further establish \benchmark{} to rigorously evaluate performance across diverse visual in-context editing tasks. We provide further details regarding the dataset and benchmark in Appendices~\ref{suppl:dataset} and~\ref{suppl:bench}.

\begin{figure}[t]
    \centering
    \includegraphics[width=0.9\linewidth]{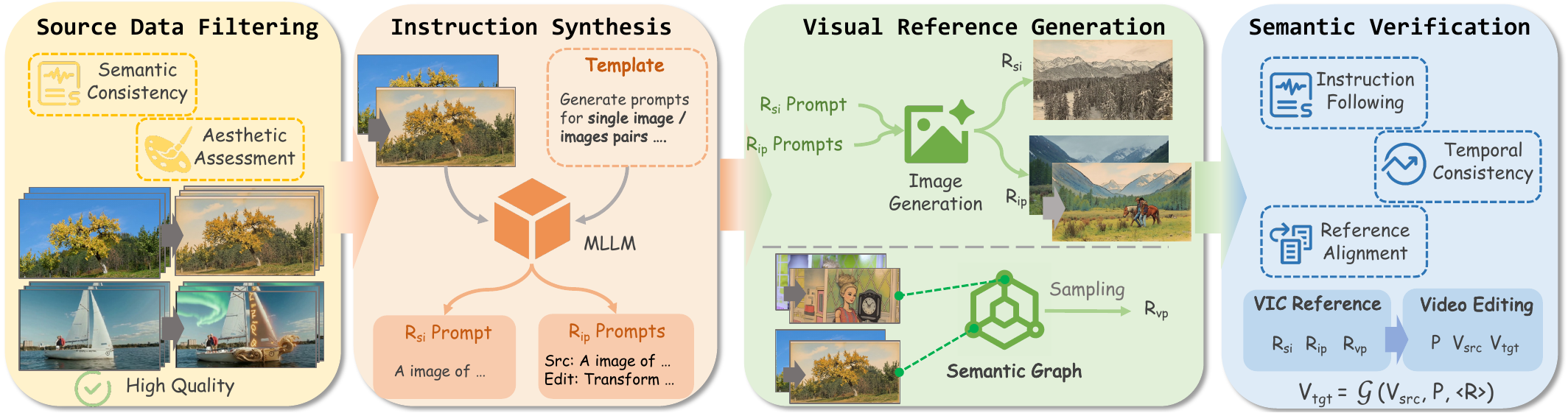}
    \caption{\textbf{Data construction pipeline} that comprises four stages (Sec.~\ref{sec:data}), 
    % Source Data Filtering, Instruction Synthesis, Visual Reference Generation, and Semantic Verification. 
    % This process 
    yielding a final dataset of 400K samples that covers conventional instruction-based editing and visual in-context editing tasks. All examples presented in the figure are collected from the open-source OpenVE~\cite{he2025openve}.}
    \label{fig:data_pipeline}
    \vspace{-3mm}
\end{figure}

\begin{figure}[t]
    \centering
    \includegraphics[width=0.9\linewidth]{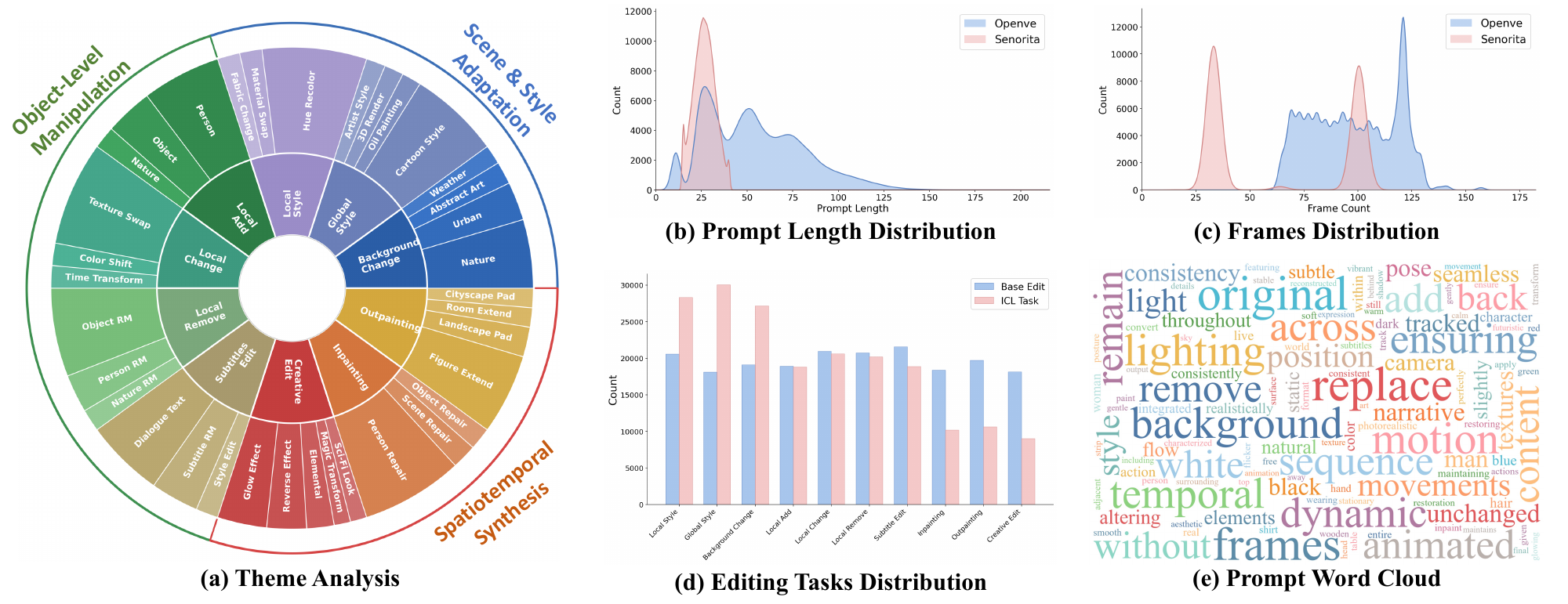}
    \caption{Statistics of \dataset{}.}
    \label{fig:data_type}
    \vspace{-5mm}
\end{figure}

\textbf{Data Source.} We build \dataset{} upon two recent large-scale datasets: Señorita-2M~\cite{zisenorita}, which covers 18 editing sub-categories, and OpenVE-3M~\cite{he2025openve}, containing approximately 3 million samples across 8 spatial and non-spatial categories. While extensive, both datasets rely solely on textual instructions and lack visual references. To address this, we systematically consolidate their editing tasks and curate a subset to construct multi-modal visual references. 

\textbf{Data Construction Pipeline.} As shown in Fig. \ref{fig:data_pipeline}, our construction pipeline comprises four stages:

\hspace*{1em} \noindent \textit{\textbf{1) Source Data Filtering.}}We filter raw data for semantic consistency and aesthetic quality~\cite{huang2024vbench,zheng2025vbench}. By leveraging the LAION-Aesthetic~\cite{schuhmann2022laion} predictor and Qwen3-VL~\cite{yang2025qwen3}, we implement a dual-model filtering mechanism that selects only those samples possessing both superior visual appeal and high semantic editability for the downstream pipeline.

\hspace*{1em} \noindent \textit{\textbf{2) Instruction Synthesis.}}
Leveraging MLLMs~\cite{bai2025qwen3-vl,yang2025qwen3}, we synthesize modality-aware instructions tailored to each reference type. By employing structured prompting, we ensure the instructions maintain strict semantic alignment with the visual references while explicitly prescribing non-trivial edits that differentiate the target from the source content. This approach yields a diverse prompt set spanning 10 core tasks, ensuring both consistency and editability.

\hspace*{1em} \noindent \textit{\textbf{3) Visual Reference Generation.}} 
We employ modality-specific strategies for reference synthesis. For single-image and image-pair references, we utilize state-of-the-art generative models~\cite{flux-2-2025,wu2025qwenimagetechnicalreport} to synthesize target content guided by the generated instructions. For video-pair references, to circumvent temporal artifacts in generative editing, we construct a semantic knowledge graph and partition the task space via spectral clustering~\cite{ von2007tutorial}. By sampling semantically-linked pairs from these clusters, we provide authentic references that maintain high visual fidelity.

\hspace*{1em} \noindent \textit{\textbf{4) Semantic Verification.}}
Final samples undergo MLLM-based multi-dimensional verification. By employing pre-defined scoring templates, we prompt MLLMs to rigorously assess instruction following, temporal consistency~\cite{lin2026kiwi,he2025openve}, and reference alignment. This standardized evaluation allows us to filter out inconsistent samples and ensure the high reliability of the final dataset.

\textbf{Data Statistics and Analysis. } \dataset{} covers three categories and 10 fine-grained tasks as shown in Fig.~\ref{fig:data_type}. Statistical analysis reveals significant diversity, with video lengths primarily concentrated in the 30$\sim$120 frame intervals. This bimodal distribution, coupled with a broad semantic prompt space (Fig.~\ref{fig:data_type}e), provides a robust foundation for advancing visual in-context editing.

\textbf{Benchmark and Evaluation.} We evaluate \method{} on two paradigms: \textit{1)} Instruction-based Editing, where we introduce OpenVE-Ext by augmenting OpenVE-Bench~\cite{he2025openve} with 3 additional tasks from Señorita~\cite{zisenorita}; and \textit{2)} Visual In-Context Editing, using our proposed \benchmark{} to assess heterogeneous reference following. \benchmark{} consists of 200 representative instances (10 per task-modality combination) strictly isolated from the training set for held-out evaluation. Following~\cite{he2025openve}, we leverage VLMs~\cite{yang2025qwen3,comanici2025gemini} for automated scoring, supplemented by fidelity and temporal metrics from VBench~\cite{huang2024vbench,zheng2025vbench} and IVE-bench~\cite{chen2025ivebench}.

% 强调一下和之前框架的不同点：之前框架不能适应新任务
\subsection{Methodology: Visual In-context Editing (VicEdit)}
\label{sec:methodology}
Existing text-only video editing frameworks are inherently task-limited and fail to generalize to complex editing scenarios requiring fine-grained spatial and semantic constraints. They cannot adapt to emerging in-context editing paradigms that rely on multi-modal visual guidance. Therefore, we introduce \method{}, a unified framework for visual in-context video editing that extends video editing to accommodate multi-modal visual guidance.

\begin{figure}
    \centering
    \includegraphics[width=\linewidth]{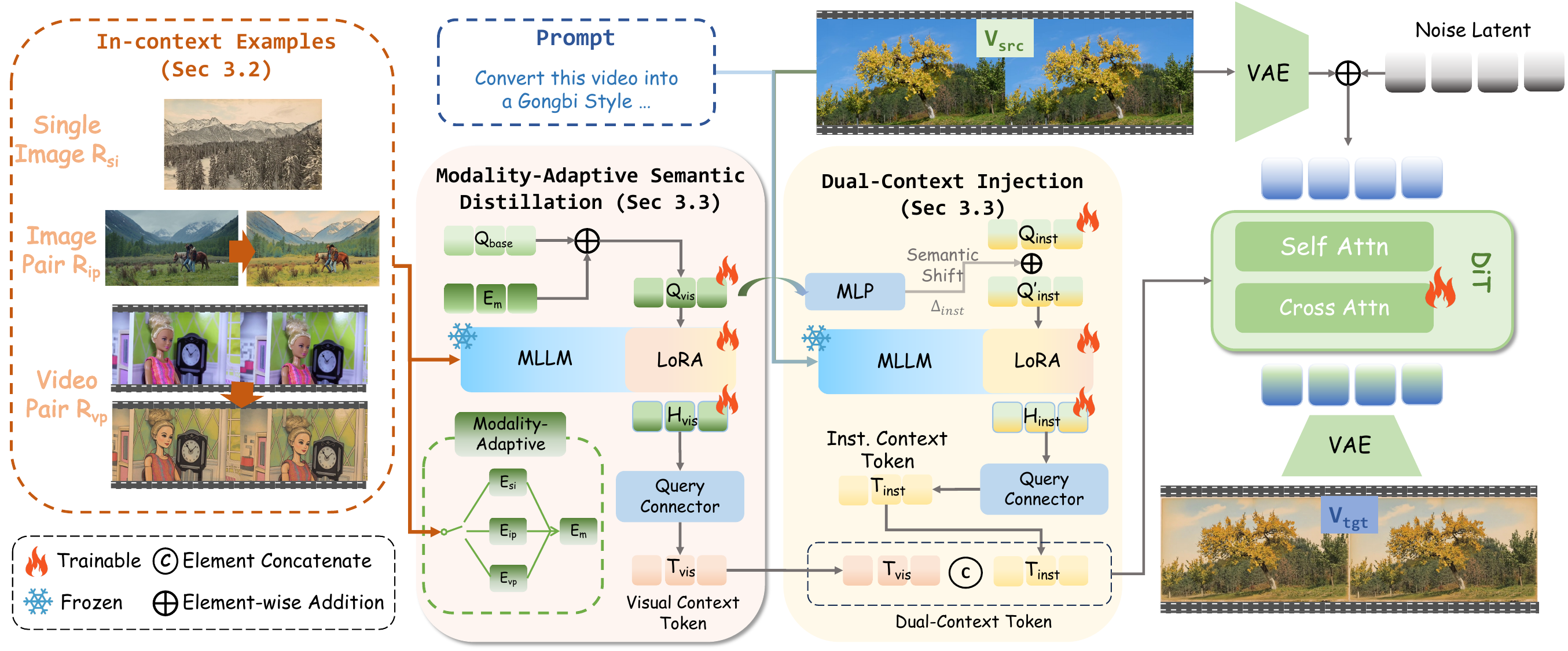}
    \caption{\textbf{Overview of \method{}.}
\textbf{\textit{1)} MASD} modulates base queries $Q_{base}$ with modality embeddings $E_m$ to distill visual context tokens $T_{vis}$.
\textbf{\textit{2)} DCI} applies a modality-conditioned shift to instruction queries.
The resulting visual and instruction context tokens are concatenated and injected into the video DiT to guide generation.}
    \label{fig:pipeline}
\end{figure}

\textbf{Overview of \method{}.} As illustrated in Figure~\ref{fig:pipeline}, \method{} employs an MLLM for semantic understanding and a DiT for generation, connected by two lightweight modules: \textbf{MASD}, which extracts modality-aware visual context tokens $\mathbf{T}_{\text{vis}}$ from the in-context reference, and \textbf{DCI}, which produces instruction context tokens $\mathbf{T}_{\text{inst}}$ using modality-conditioned instruction queries. $\mathbf{T}_{\text{vis}}$ and $\mathbf{T}_{\text{inst}}$ encode complementary perceptual editing semantics and task-level edit directions. These token sequences are concatenated as cross-attention keys and values for the DiT to generate $\mathbf{V}_e$.

\textbf{Modality-Adaptive Semantic Distillation.} 
The three reference modalities capture structurally distinct editing semantics: image pairs reflect spatial transformations, video pairs provide temporal motion priors, and single images anchor global appearance attributes. Existing methods~\cite{tan2025omnivideo,wei2025univideo,lin2026kiwi} often extract semantic tokens using a fixed set of learnable queries, implicitly assuming a universal pattern for all visual cues. However, this assumption fails under heterogeneous modalities, as a shared query cannot adaptively distinguish between the localized spatial shifts in an image pair versus the global stylistic textures in a single reference image.

To explicitly account for reference modality, MASD introduces an additive, modality-conditioned offset:
\begin{equation}
\mathbf{Q}_{\text{vis}}
= \mathbf{Q}_{\text{base}} + \mathbf{E}_m,
\quad m \in \{\text{ip}, \text{vp}, \text{si}\},
\label{eq:q_vis}
\end{equation}
where $\mathbf{Q}_{\text{base}} \in \mathbb{R}^{N \times d}$ denotes learnable queries shared across reference modalities, and $\mathbf{E}_m \in \mathbb{R}^{N \times d}$ is a learnable modality-specific embedding. This parameterization combines shared queries with modality-dependent offsets, allowing the extraction to adapt to heterogeneous reference types.

We append $\mathbf{Q}_{\text{vis}}$ to the reference input and process the sequence through the MLLM. The output hidden states at the query positions, denoted by $\mathbf{H}_{\text{vis}}$, are projected into $N$ visual context tokens through an MLP
Query Connector:
\begin{equation}
\mathbf{T}_{\text{vis}}
= \mathrm{MLP}_{\text{vis}}(\mathbf{H}_{\text{vis}}).
\label{eq:t_vis}
\end{equation}
The connector maps the query hidden states into the DiT conditioning space. While the query embeddings depend on the reference modality, the resulting tokens also capture instance-specific visual content.
The shared queries and modality embeddings are learned under the editing objective.

\textbf{Dual-Context Injection.}
DCI couples visual and instruction extraction through their query
parameterizations. Specifically, it maps the modality-conditioned
visual queries to an instruction-query shift:
\begin{equation}
\boldsymbol{\Delta}_{\text{inst}}
= \mathrm{MLP}_{\text{shift}}(\mathbf{Q}_{\text{vis}})
\in \mathbb{R}^{N \times d},
\label{eq:shift}
\end{equation}
which is added to the learnable instruction queries:
\begin{equation}
\mathbf{Q}_{\text{inst}}'
= \mathbf{Q}_{\text{inst}} + \boldsymbol{\Delta}_{\text{inst}}.
\label{eq:q_inst}
\end{equation}
Since $\mathbf{Q}_{\text{vis}}=\mathbf{Q}_{\text{base}}+\mathbf{E}_m$,
the shift depends on the reference modality and is shared across
instances of the same modality at inference. Without MASD, it reduces
to a global learned shift. During training, the shared query parameters
receive gradients through both visual extraction and instruction-query
modulation, coupling the two branches.

We append $\mathbf{Q}_{\text{inst}}'$ to the text prompt and process
the sequence through the MLLM. The output hidden states at the query
positions, denoted by $\mathbf{H}_{\text{inst}}$, are projected into
instruction context tokens through an MLP Query Connector:
\begin{equation}
\mathbf{T}_{\text{inst}}
= \mathrm{MLP}_{\text{inst}}(\mathbf{H}_{\text{inst}}).
\label{eq:t_inst}
\end{equation}

Finally, we concatenate the two token sequences as
$\mathbf{T}_{\text{dual}}
=[\mathbf{T}_{\text{vis}};\mathbf{T}_{\text{inst}}]$
and inject them as keys and values into the video DiT cross-attention
layers. This supplies the generator with instance-specific visual
evidence and modality-conditioned instruction representations.

\textbf{Training Strategy.} \method{} builds upon Wan-TI2V-5B~\cite{wan2025wan} and inherits pretrained weights from Kiwi-Edit. \method{} is trained in three stages to progressively incorporate in-context editing capability.

\hspace*{1em} \noindent \textit{\textbf{1) Base Editing.}} We first fine-tune the pretrained video DiT on instruction-based editing tasks to expand its editing capabilities while promoting instruction following and temporal consistency. The MLLM is initialized from pretrained weights and kept frozen.

\hspace*{1em} \noindent \textit{\textbf{2) In-Context Branch.}} With the DiT frozen, we train the in-context modules, including MASD, the semantic shift MLP in DCI, and the instruction queries $\mathbf{Q}_{\text{inst}}$, together with rank-64 LoRA adapters in the MLLM language component. Each training batch mixes samples from all three modalities to ensure balanced learning of the modality embeddings. 

\hspace*{1em} \noindent \textit{\textbf{3) Joint Fine-Tuning.}} We unfreeze the DiT and jointly fine-tune the entire pipeline (except the MLLM) to align the visual conditioning signals with the generative process. All stages are optimized with the standard denoising loss: \begin{equation} 
\mathcal{L} = \mathbb{E}_{\mathbf{V}_e, \boldsymbol{\epsilon}, t} \bigl[ \| \boldsymbol{\epsilon} - \boldsymbol{\epsilon}_\theta(\mathbf{V}_e^{(t)}, t, \mathbf{T}_{\text{dual}}) \|^2 \bigr], 
\label{eq:loss} 
\end{equation} 
where $\mathbf{V}_e^{(t)}$ is the noised edited video at timestep $t$, $\boldsymbol{\epsilon}$ is the ground-truth noise, and $\boldsymbol{\epsilon}_\theta$ denotes the DiT noise predictor conditioned on $\mathbf{T}_{\text{dual}}$.

\section{Experiments}

% 表头不用突出Gemini，可以点出base editing
\begin{table}[t]
\centering
\caption{Quantitative Comparison on OpenVE-Ext with Gemini 2.5 Pro~\cite{comanici2025gemini}.}
\label{tab:openve_bench}
\resizebox{\textwidth}{!}{%
\begin{tabular}{l|c|ccc|cccc|ccc}
\toprule
 & & \multicolumn{3}{c|}{\textbf{Scene \& Style Adaptation}} & \multicolumn{4}{c|}{\textbf{Object-Level Manipulation}} & \multicolumn{3}{c}{\textbf{Spatiotemporal Synthesis}} \\
\cmidrule(lr){3-5} \cmidrule(lr){6-9} \cmidrule(lr){10-12}
\textbf{Methods} 
& \textbf{Overall} & \makecell[c]{\textbf{Global}\\\textbf{Style}} & \makecell[c]{\textbf{Local}\\\textbf{Style}} & \makecell[c]{\textbf{Background}\\\textbf{Change}} 
& \makecell[c]{\textbf{Local}\\\textbf{Change}} & \makecell[c]{\textbf{Local}\\\textbf{Remove}} & \makecell[c]{\textbf{Local}\\\textbf{Add}} 
& \makecell[c]{\textbf{Subtitle}\\\textbf{Edit}} & \makecell[c]{\textbf{Creative}\\\textbf{Edit}} & \textbf{Inpainting} & \textbf{Outpainting} \\
\midrule
VACE~\cite{jiang2025vace} & 1.68 & 1.49 & 1.39 & 1.55 & 2.07 & 1.46 & 1.26 & 1.48 & 1.47 & 2.31 & 2.35 \\
Lucy-Edit~\cite{team2025lucy} & 2.35 & 2.27 & 2.02 & 1.57 & 3.20 & 1.75 & 2.30 & 1.61 & 2.86 & 3.01 & 2.92 \\
DITTO~\cite{bai2025ditto} & 2.31 & \textbf{4.01} & \textbf{3.67} & 1.68 & 2.03 & 1.53 & 1.41 & 2.81 & 1.23 & 2.05 & 2.67 \\
NovaEdit~\cite{pan2026nova} & 2.76 & 2.24 & 1.98 & 2.40 & 3.62 & 3.43 & 2.64 & 2.93 & 2.59 & 2.78 & 3.01 \\
VideoCoF~\cite{yang2026videocof} & 2.79 & 2.57 & 2.24 & 2.32 & 3.37 & 3.61 & \textbf{2.75} & 3.10 & 2.20 & 2.87 & 2.88 \\
Kiwi-Edit~\cite{lin2026kiwi} & 3.08 & 3.64 & 2.85 & 2.64 & 3.83 & 2.63 & 2.36 & 3.20 & \textbf{3.39} & 3.35 & 2.86 \\
\midrule 
\rowcolor{cyan!15}
\textbf{VicEdit (Ours)} & \textbf{3.28} & 3.48 & 2.90 & \textbf{3.05} & \textbf{3.85} & \textbf{3.68} & 2.64 & \textbf{3.32} & 3.25 & \textbf{3.38} & \textbf{3.20} \\
\bottomrule
\end{tabular}
}
\end{table}
\begin{figure}[t]
    \centering
    \includegraphics[width=0.92\linewidth]{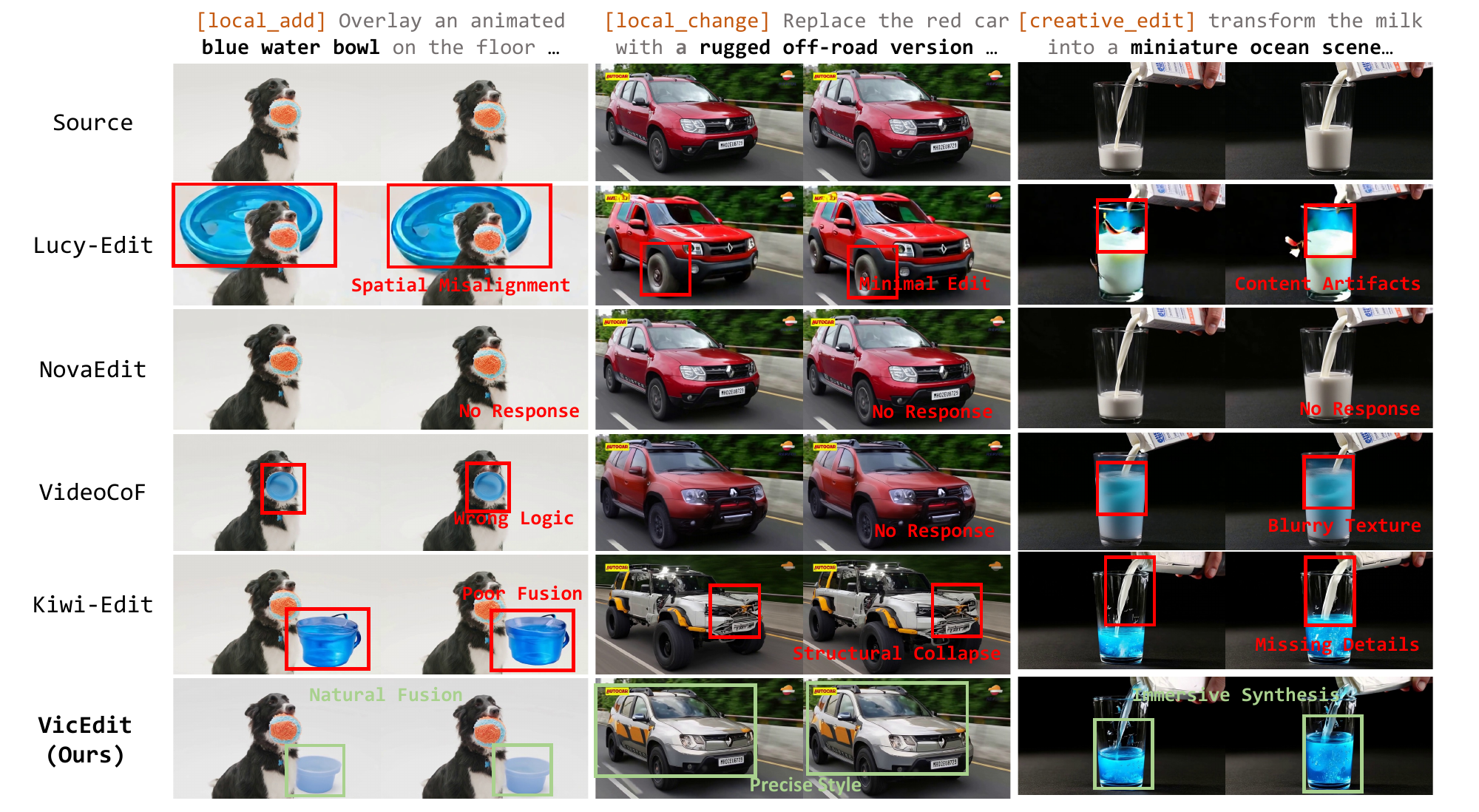}
    \caption{Comparison with baselines on instruction-based editing tasks.}
    \label{fig:main_exp}
\end{figure}

% 表头不用突出Gemini，可以点出visual in-context editing
\begin{table}[t]
\centering
\caption{Quantitative comparison on \benchmark{} with Gemini 2.5 Pro~\cite{comanici2025gemini}.}
\label{tab:our_bench}
\resizebox{\textwidth}{!}{%
\begin{tabular}{l|c|ccc|cccc|ccc}
\toprule
 & & \multicolumn{3}{c|}{\textbf{Scene \& Style Adaptation}} & \multicolumn{4}{c|}{\textbf{Object-Level Manipulation}} & \multicolumn{3}{c}{\textbf{Spatiotemporal Synthesis}} \\
\cmidrule(lr){3-5} \cmidrule(lr){6-9} \cmidrule(lr){10-12}
\textbf{Methods} 
& \textbf{Overall} & \makecell[c]{\textbf{Global}\\\textbf{Style}} & \makecell[c]{\textbf{Local}\\\textbf{Style}} & \makecell[c]{\textbf{Background}\\\textbf{Change}} 
& \makecell[c]{\textbf{Local}\\\textbf{Change}} & \makecell[c]{\textbf{Local}\\\textbf{Remove}} & \makecell[c]{\textbf{Local}\\\textbf{Add}} 
& \makecell[c]{\textbf{Subtitle}\\\textbf{Edit}} & \makecell[c]{\textbf{Creative}\\\textbf{Edit}} & \textbf{Inpainting} & \textbf{Outpainting} \\
\midrule
VACE~\cite{jiang2025vace} & 1.85 & 1.45 & 2.03 & 1.50 & 2.25 & 1.49 & 1.33 & 1.43 & 2.17 & 2.23 & 2.60 \\
Lucy-Edit~\cite{team2025lucy} & 2.25 & 1.94 & 1.67 & 1.89 & 3.30 & 2.05 & 2.15 & 1.43 & 2.45 & 2.63 & 2.97\\
DITTO~\cite{bai2025ditto} & 2.33 & 3.03 & \textbf{3.60} & 2.29 & 2.45 & 1.64 & 1.43 & 2.61 & 1.46 & 2.48 & 2.32\\
NovaEdit~\cite{pan2026nova} & 2.61 & 2.08 & 2.23 & 2.62 & 3.07 & 3.65 & 2.05 & 2.46 & 2.13 & 2.65 & 3.18 \\
VideoCoF~\cite{yang2026videocof} & 2.90 & 2.66 & 2.43 & 2.37 & 3.67 & 3.60 & 2.44 & \textbf{3.65} & 1.83 & 3.06 & 3.33   \\
Kiwi-Edit~\cite{lin2026kiwi} & 2.95 & 3.19 & 2.39 & 2.53 & \textbf{4.29} & 3.24 & 2.28 & 3.16 & 2.77 & 3.18 & 2.56 \\
\midrule 
\rowcolor{cyan!15}
\textbf{VicEdit (Ours)} & \textbf{3.45} & \textbf{3.68} & 3.03 & \textbf{3.05} & 4.17 & \textbf{3.86} & \textbf{2.85} & 3.62 & \textbf{3.30} & \textbf{3.45} & \textbf{3.47} \\
\bottomrule
\end{tabular}
}
\end{table}

\begin{figure}
    \centering
    \includegraphics[width=0.92\linewidth]{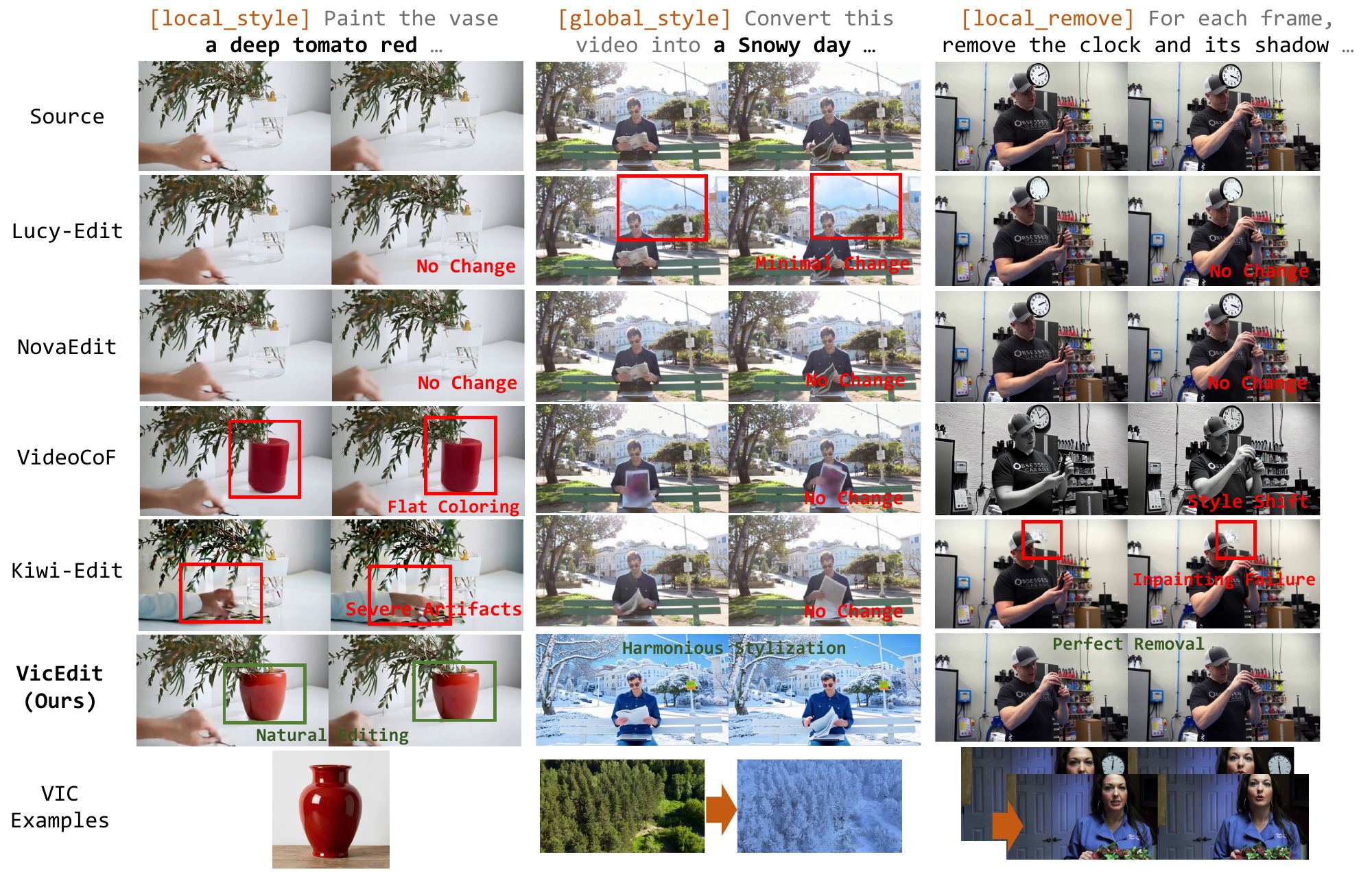}
    \caption{Comparison with baselines on visual in-context editing tasks. All examples are from OpenVE~\cite{he2025openve}.}
    \label{fig:main_vic}
\end{figure}

\subsection{Base Video Editing}
\textbf{Quantitative Analysis.} We compare VicEdit with six video editing methods~\cite{jiang2025vace,team2025lucy,bai2025ditto,pan2026nova,yang2026videocof,lin2026kiwi} on OpenVE-Ext. Table~\ref{tab:openve_bench} reports an Overall score of 3.28 for VicEdit and 3.08 for Kiwi-Edit. Performance varies by task: for example, DITTO is strong on global and local style, while VicEdit achieves the highest reported scores on background change, local change, local removal, subtitle editing, inpainting, and outpainting. These text-only results characterize the trained system's editing ability; they do not independently isolate the contributions of the visual context modules. Component ablations are reported in Table~\ref{tab:ablation_full}.

\textbf{Qualitative Analysis.} Fig. \ref{fig:main_exp} presents several qualitative examples. As shown in the comparisons, our method achieves superior performance in spatial layout preservation, semantic comprehension, and visual quality. For instance, in the ``local\_add'' task, baselines struggle to handle the spatial relationships between the newly added object and existing objects in terms of position and scale, whereas \method{} produces well-structured editing results. In the ``local\_change'' task, most baselines fail to understand the semantic transformation, and Kiwi-Edit, while demonstrating semantic understanding, introduces noticeable artifacts. In the ``creative\_edit'' example, videos generated by our method are more semantically coherent and logically consistent with the given prompts.

\subsection{Visual In-Context Video Editing}
\textbf{Quantitative Analysis.} To comprehensively evaluate the effectiveness of our proposed task and model, we conduct experiments on the \benchmark{} benchmark. Crucially, unlike prior works that rely solely on textual instructions, our evaluation setting incorporates visual in-context examples as additional references, which is the key innovation of our VIC task. For a fair comparison, we equip all baselines with an instruction enhancement module to augment their text-only instructions. As shown in Table \ref{tab:our_bench}, VicEdit achieves a remarkable Overall score of 3.45. This performance surpasses the second-best method, Kiwi-Edit, by a substantial margin of 0.50 points (a 17\% relative improvement). The significant gap underscores VicEdit's superior capability in understanding and executing complex editing instructions. In contrast, previous text-only approaches often struggle to capture precise semantics from text descriptions alone, leading to suboptimal editing results.

\textbf{Qualitative Analysis.}
Qualitative results on the visual in-context task are shown in Figure~\ref{fig:main_vic}. 
A key observation is that our model, conditioned on diverse visual contexts (single images, image pair,  video pair), effectively captures the instruction semantics. 
Taking the \textbf{Global Style} task as an example, converting the scene to a snowy day proves challenging for existing text-only baselines, which struggle to synthesize a visually plausible snowy style. However, equipped with visual in-context example, our model seamlessly adapts to the desired style, highlighting its robust editing capability.

\subsection{Human Evaluation}
\begin{table}[t]
\centering
\caption{Human preferences and evaluator agreement.
Ten annotators each evaluate 20 examples per setting.
Preference scores compare against the strongest task-specific baseline, awarding 1 point for wins and 0.5 for ties.
Gemini--Qwen correlation measures judge--judge agreement separately.}
\label{tab:human_eval}
\small
\setlength{\tabcolsep}{5pt}
\begin{tabular}{lcc}
\toprule
Metric & Base Editing & VIC Editing \\
\midrule
Human Preference Score (\%) $\uparrow$ & 76 & 72 \\
Human--Gemini Agreement (\%) $\uparrow$ & 80 & 76 \\
Gemini--Qwen-VL Pearson's $r$ $\uparrow$ & 0.76 & 0.74 \\
Fleiss' $\kappa$ $\uparrow$ & 0.70 & 0.66 \\
\bottomrule
\end{tabular}
\end{table}

We conduct pairwise human preference studies on base editing and visual in-context editing, comparing VicEdit with the strongest baseline for each selected task, with ten annotators each evaluating 20 examples per setting. As shown in Table~\ref{tab:human_eval}, VicEdit achieves tie-adjusted preference scores of 76\% and 72\%, respectively, with each tie contributing half a point to each method. Human judgments agree with Gemini 2.5 Pro at rates of 80\% and 76\%, while inter-annotator agreement reaches Fleiss' $\kappa$ values of 0.70 and 0.66, respectively. Separately, Gemini 2.5 Pro and Qwen-VL scores have Pearson correlations of 0.76 and 0.74. These results provide complementary human evidence for VicEdit's effectiveness; the small study does not establish general reliability or judge independence of VLM scoring. Appendix~\ref{suppl:evaluation} describes the evaluation protocol.

\subsection{Ablation Study}
\begin{table}[t]
\centering
\caption{Component ablation on \benchmark{}. VLM scores are mean $\pm$ standard deviation over three video-generation runs, using one fixed trained checkpoint per configuration. Shared-Query Base uses source-video and text conditioning without visual-reference tokens. The other configurations incorporate visual-reference tokens; Naive Concatenation combines them with instruction tokens by direct concatenation without MASD or DCI.}
\label{tab:ablation_full}
\resizebox{\linewidth}{!}{%
\begin{tabular}{l|cccc|ccccc}
\toprule
 & \multicolumn{4}{c|}{\textbf{VLM Evaluation}} & \multicolumn{5}{c}{\textbf{VBench Metrics}} \\
\cmidrule(lr){2-5}\cmidrule(lr){6-10}
Variant & Overall$\uparrow$ & \makecell{Instr.\\Comp.$\uparrow$} & \makecell{Detail\\Fid.$\uparrow$} & \makecell{Visual\\Qual.$\uparrow$} & BC$\uparrow$ & MS$\uparrow$ & TF$\uparrow$ & IQ$\uparrow$ & AQ$\uparrow$ \\
\midrule
Shared-Query Base & $3.02\pm0.05$ & $3.12\pm0.06$ & $2.98\pm0.06$ & $2.96\pm0.05$ & 0.968 & 0.995 & 0.986 & 0.711 & 0.564 \\
Naive Concatenation & $3.14\pm0.05$ & $3.24\pm0.05$ & $3.09\pm0.05$ & $3.09\pm0.06$ & 0.970 & 0.993 & 0.985 & 0.712 & 0.566 \\
MASD only & $3.25\pm0.04$ & $3.40\pm0.04$ & $3.20\pm0.05$ & $3.14\pm0.05$ & 0.969 & 0.994 & 0.986 & 0.712 & 0.566 \\
DCI only & $3.34\pm0.04$ & $3.56\pm0.03$ & $3.22\pm0.04$ & $3.24\pm0.04$ & 0.971 & 0.995 & 0.986 & 0.715 & 0.567 \\
\midrule
\rowcolor{cyan!10}
MASD + DCI & $\mathbf{3.45\pm0.03}$ & $3.65\pm0.03$ & $3.34\pm0.04$ & $3.35\pm0.03$ & \textbf{0.975} & 0.994 & \textbf{0.987} & \textbf{0.718} & \textbf{0.568} \\
\bottomrule
\end{tabular}}
\end{table}

\textbf{Ablation configurations.}
Table~\ref{tab:ablation_full} compares five configurations. Shared-Query Base is a Kiwi-Edit-style fixed-query model fine-tuned on VicEdit-400K, conditioned on the source video
and instruction without visual-reference tokens. Naive Concatenation adds visual-reference tokens using separate visual and instruction queries, concatenating
$[\mathbf{T}_{\text{vis}};\mathbf{T}_{\text{inst}}]$ for DiT conditioning without modality embeddings or DCI. MASD only adds modality embeddings to this baseline;
DCI only adds query-shift coupling with
$\mathbf{Q}_{\text{vis}}=\mathbf{Q}_{\text{base}}$. The full model combines both modules.

\textbf{Ablation on MASD.}
Table~\ref{tab:ablation_full} evaluates modality-conditioned queries under two configurations. Without DCI, adding MASD increases Overall from 3.14 to 3.25. With DCI enabled, adding MASD increases Overall from 3.34 to 3.45 and detail fidelity from 3.22 to 3.34. These results support the effectiveness of modality-specific query offsets for heterogeneous reference extraction under the evaluated setup. They do not establish an explicitly enforced decomposition of semantic and spatial information.

\textbf{Ablation on DCI.}
DCI improves Overall from 3.14 to 3.34 without MASD, and from 3.25 to 3.45 with MASD. Without MASD, the shift $\mathrm{MLP}(\mathbf{Q}_{\text{base}})$ is reference-independent at inference and can be absorbed into freely learned instruction queries. During training, however, DCI couples visual extraction and instruction-query modulation through the shared query parameters. With MASD, the shift additionally depends on the reference modality, while instance-specific visual content is conveyed through $\mathbf{T}_{\text{vis}}$. The results support the effectiveness of this parameterization, although the precise optimization mechanism responsible for the gains is not isolated by these ablations.

The synergistic integration of both modules achieves the best overall performance, with detailed visual comparisons provided in Appendix~\ref{suppl:visual_ablation}.

\begin{table}[!t]
\centering
\caption{Quantitative Ablation on Training Stages. We evaluate the performance evolution across different training phases on the OpenVE-Ext and \benchmark{} benchmarks.}
\label{tab:ablation_training_stages}
\small
\setlength{\tabcolsep}{3.5pt} 
\begin{tabular}{l|cccc|cccc}
\toprule
 & \multicolumn{4}{c|}{\textbf{Base Editing}} & \multicolumn{4}{c}{\textbf{Visual In-Context Editing}} \\
\cmidrule(lr){2-5} \cmidrule(lr){6-9}
\textbf{Training Stage} & Overall$\uparrow$ & \makecell[c]{Instr.\\Comp.$\uparrow$} & \makecell[c]{Detail\\Fid.$\uparrow$} & \makecell[c]{Visual\\Qual.$\uparrow$} & Overall$\uparrow$ & \makecell[c]{Instr.\\Comp.$\uparrow$} & \makecell[c]{Detail\\Fid.$\uparrow$} & \makecell[c]{Visual\\Qual.$\uparrow$} \\
\midrule
Base (Kiwi-Edit) & 3.08 & 3.24 & 3.01 & 2.98 & 2.95 & 3.10 & 2.98 & 2.76  \\
Stage 1 & 3.23 & 3.36 & 3.11 & 3.20 & 3.10 & 3.25 & 3.05 & 3.01 \\
Stage 2 & 3.10 & 3.22 & 3.02 & 3.07 & 3.40 & 3.62 & 3.30 & 3.27 \\
Stage 3 & \textbf{3.28} & 3.47 & 3.27 & 3.10  & \textbf{3.45} & 3.65 & 3.34 & 3.35 \\
\bottomrule
\end{tabular}
\end{table}
\begin{table}[!t]
\centering
\caption{Separating the Contributions of Training Data and Model Architecture.}
\label{tab:data_arch_ablation}
\small
\setlength{\tabcolsep}{5pt}
\begin{tabular}{lcccc}
\toprule
Method & Overall $\uparrow$ & Instr. Comp. $\uparrow$ & Detail Fid. $\uparrow$ & Visual Qual. $\uparrow$ \\
\midrule
Kiwi-Edit (original) & 2.95 & 3.10 & 2.98 & 2.76 \\
Kiwi-style + VicEdit-400K & 3.14 & 3.24 & 3.09 & 3.09 \\
VicEdit (full) & \textbf{3.45} & \textbf{3.65} & \textbf{3.34} & \textbf{3.35} \\
\bottomrule
\end{tabular}
\end{table}

\begin{table}[!t]
\centering
\caption{Ablation on Visual Reference Effectiveness. We compare correct, mismatched, and absent visual references to evaluate their impact on editing performance and reference adherence.}
\label{tab:visual_reference_effectiveness}
\small
\setlength{\tabcolsep}{5pt}
\begin{tabular}{lcccc}
\toprule
Setting & Instr. Comp. $\uparrow$ & Detail Fid. $\uparrow$ & Visual Qual. $\uparrow$ & Ref. Adherence $\uparrow$ \\
\midrule
Text-only (no reference) & 3.12 & 3.08 & 3.10 & -- \\
Mismatched reference & 2.90 & 2.82 & 2.68 & 1.68 \\
Correct reference & \textbf{3.58} & \textbf{3.28} & \textbf{3.28} & \textbf{3.42} \\
\bottomrule
\end{tabular}
\end{table}

\textbf{Ablation on Training Strategy.}
As shown in Table~\ref{tab:ablation_training_stages},
Stage 1 broadens the range of editing tasks, improving both base and visual in-context editing across all metrics. Stage 2 focuses on visual in-context learning, further improving VIC performance while reducing base editing performance. Finally, Stage 3 jointly trains both tasks, recovering base editing performance and further improving VIC editing,
yielding the highest overall scores in both settings.

\textbf{Ablation on Dataset and Architecture Contributions.} To disentangle the contributions of VicEdit-400K and our proposed architecture, we compare the original Kiwi-Edit, a Kiwi-style variant trained on VicEdit-400K, and the full VicEdit model. Replacing the original training data with VicEdit-400K improves the overall score from 2.95 to 3.14, demonstrating the benefit of the proposed dataset. With the training data and experimental setup held constant, incorporating MASD and DCI further increases the score to 3.45. The full model also achieves consistent gains in instruction compliance, detail fidelity, and visual quality. These results show that the improvement comes from both the dataset and the proposed architecture, with their combination yielding the strongest overall performance.

\textbf{Ablation on Visual Reference Effectiveness.} To assess whether VicEdit uses visual references beyond textual instructions, we compare correct, mismatched, and no-reference settings. Correct references improve instruction compliance, detail fidelity, and visual quality over text-only conditioning, whereas mismatched references reduce
all three metrics below the text-only baseline.
Reference adherence also drops substantially with mismatched references.
This pattern indicates that visual references actively influence the editing process: relevant references provide useful guidance, while misleading ones can interfere
with instruction execution. The benefits of visual conditioning therefore depend on reference relevance.
\section{Conclusion}
\label{conclusion}
We present \method{}, a unified framework extending text-only video editing to multi-modal visual guidance. Modality-Adaptive Semantic Distillation (MASD) extracts editing semantics through modality-conditioned learnable queries, while Dual-Context Injection (DCI) couples visual and textual contexts through a semantic shift, enabling fine-grained control across single images, image pairs, and video pairs. We construct \dataset{}, the first large-scale dataset for visual in-context video editing, with 400K samples across ten tasks, and introduce \benchmark{} for standardized evaluation. Extensive experiments demonstrate state-of-the-art performance on both instruction-based and visual in-context editing benchmarks, establishing a scalable paradigm for video editing. Our current design targets semantic-level editing; future work may explore pixel-level control and larger-scale datasets.

\section*{Acknowledgements}
This work was supported by the National Natural Science Foundation of China (No. 72192821, 62302297, 62472282, 62272447, 62472285, 625B2115).

\bibliography{main}

@inproceedings{wu2023tuneavideo,
  title={Tune-a-video: One-shot tuning of image diffusion models for text-to-video generation},
  author={Wu, Jay Zhangjie and Ge, Yixiao and Wang, Xintao and Lei, Stan Weixian and Gu, Yuchao and Shi, Yufei and Hsu, Wynne and Shan, Ying and Qie, Xiaohu and Shou, Mike Zheng},
  booktitle={Proceedings of the IEEE/CVF international conference on computer vision},
  pages={7623--7633},
  year={2023}
}

@article{geyer2023tokenflow,
  title={Tokenflow: Consistent diffusion features for consistent video editing},
  author={Geyer, Michal and Bar-Tal, Omer and Bagon, Shai and Dekel, Tali},
  journal={arXiv preprint arXiv:2307.10373},
  year={2023}
}

@article{ku2024anyv2v,
  title={Anyv2v: A tuning-free framework for any video-to-video editing tasks},
  author={Ku, Max and Wei, Cong and Ren, Weiming and Yang, Harry and Chen, Wenhu},
  journal={arXiv preprint arXiv:2403.14468},
  year={2024}
}

@article{sun2024survey,
  title={Diffusion model-based video editing: A survey},
  author={Sun, Wenhao and Tu, Rong-Cheng and Liao, Jingyi and Tao, Dacheng},
  journal={arXiv preprint arXiv:2407.07111},
  year={2024}
}

@article{mai2025easyv2v,
  title={EasyV2V: A High-quality Instruction-based Video Editing Framework},
  author={Mai, Jinjie and Wang, Chaoyang and Qian, Guocheng Gordon and Menapace, Willi and Tulyakov, Sergey and Ghanem, Bernard and Wonka, Peter and Mirzaei, Ashkan},
  journal={arXiv preprint arXiv:2512.16920},
  year={2025}
}

@article{pan2026nova,
  title={NOVA: Sparse Control, Dense Synthesis for Pair-Free Video Editing},
  author={Pan, Tianlin and Dai, Jiayi and Yuan, Chenpu and Lv, Zhengyao and Yang, Binxin and Yin, Hubery and Li, Chen and Lyu, Jing and Shan, Caifeng and Si, Chenyang},
  journal={arXiv preprint arXiv:2603.02802},
  year={2026}
}

@article{lin2026kiwi,
  title={Kiwi-Edit: Versatile Video Editing via Instruction and Reference Guidance},
  author={Lin, Yiqi and Liang, Guoqiang and Zeng, Ziyun and Bai, Zechen and Chen, Yanzhe and Shou, Mike Zheng},
  journal={arXiv preprint arXiv:2603.02175},
  year={2026}
}

@article{yang2026videocof,
  title={Unified Video Editing with Temporal Reasoner},
  author={Yang, Xiangpeng and Xie, Ji and Yang, Yiyuan and Huang, Yan and Xu, Min and Wu, Qiang},
  journal={arXiv preprint arXiv:2512.07469},
  year={2025}
}

@inproceedings{kim2025temporal,
  title={Temporal In-Context Fine-Tuning with Temporal Reasoning for Versatile Control of Video Diffusion Models},
  author={Kim, Kinam and Hyung, Junha and Choo, Jaegul},
  booktitle={The Thirty-ninth Annual Conference on Neural Information Processing Systems},
  year={2025}
}

@article{yang2023imagebrush,
  title={Imagebrush: Learning visual in-context instructions for exemplar-based image manipulation},
  author={Yang, Yifan and Peng, Houwen and Shen, Yifei and Yang, Yuqing and Hu, Han and Qiu, Lili and Koike, Hideki and others},
  journal={Advances in Neural Information Processing Systems},
  volume={36},
  pages={48723--48743},
  year={2023}
}

@inproceedings{yang2023paint,
  title={Paint by example: Exemplar-based image editing with diffusion models},
  author={Yang, Binxin and Gu, Shuyang and Zhang, Bo and Zhang, Ting and Chen, Xuejin and Sun, Xiaoyan and Chen, Dong and Wen, Fang},
  booktitle={Proceedings of the IEEE/CVF conference on computer vision and pattern recognition},
  pages={18381--18391},
  year={2023}
}

@article{chen2025edittransfer,
  title={Edit Transfer: Learning Image Editing via Vision In-Context Relations},
  author={Chen, Lan and Mao, Qi and Gu, Yuchao and Shou, Mike Zheng},
  journal={arXiv preprint arXiv:2503.13327},
  year={2025}
}

@inproceedings{zhang2025vicl,
  title={Video In-context Learning: Autoregressive Transformers are Zero-Shot Video Imitators},
  author={Zhang, Wentao and Guo, Junliang and He, Tianyu and Zhao, Li and Xu, Linli and Bian, Jiang},
  booktitle={The Thirteenth International Conference on Learning Representations},
  year={2025}
}

@article{fei2024video,
  title={Video Diffusion Transformers are In-Context Learners},
  author={Fei, Zhengcong and Qiu, Di and Yu, Changqian and Li, Debang and Fan, Mingyuan and Wen, Xiang},
  journal={arXiv e-prints},
  pages={arXiv--2412},
  year={2024}
}

@article{lei2025unitransfer,
  title={UniTransfer: Video Concept Transfer via Progressive Spatial and Timestep Decomposition},
  author={Lei, Guojun and Zhang, Rong and Wang, Chi and Liu, Tianhang and Li, Hong and Ma, Zhiyuan and Xu, Weiwei},
  journal={arXiv preprint arXiv:2509.21086},
  year={2025}
}

@article{zhang2025reco,
  title={Region-Constraint In-Context Generation for Instructional Video Editing},
  author={Zhang, Zhongwei and Long, Fuchen and Li, Wei and Qiu, Zhaofan and Liu, Wu and Yao, Ting and Mei, Tao},
  journal={arXiv preprint arXiv:2512.17650},
  year={2025}
}

@article{zhang2026omnitransfer,
  title={OmniTransfer: All-in-one Framework for Spatio-temporal Video Transfer},
  author={Zhang, Pengze and Wu, Yanze and Li, Mengtian and Bai, Xu and Zhao, Songtao and Ye, Fulong and Mou, Chong and Li, Xinghui and Chen, Zhuowei and He, Qian and others},
  journal={arXiv preprint arXiv:2601.14250},
  year={2026}
}

@article{ju2025editverse,
  title={Editverse: Unifying image and video editing and generation with in-context learning},
  author={Ju, Xuan and Wang, Tianyu and Zhou, Yuqian and Zhang, He and Liu, Qing and Zhao, Nanxuan and Zhang, Zhifei and Li, Yijun and Cai, Yuanhao and Liu, Shaoteng and others},
  journal={arXiv preprint arXiv:2509.20360},
  year={2025}
}

@inproceedings{wang2023images,
  title={Images speak in images: A generalist painter for in-context visual learning},
  author={Wang, Xinlong and Wang, Wen and Cao, Yue and Shen, Chunhua and Huang, Tiejun},
  booktitle={Proceedings of the IEEE/CVF Conference on Computer Vision and Pattern Recognition},
  pages={6830--6839},
  year={2023}
}

@article{wang2023promptdiffusion,
  title={In-context learning unlocked for diffusion models},
  author={Wang, Zhendong and Jiang, Yifan and Lu, Yadong and He, Pengcheng and Chen, Weizhu and Wang, Zhangyang and Zhou, Mingyuan and others},
  journal={Advances in Neural Information Processing Systems},
  volume={36},
  pages={8542--8562},
  year={2023}
}

@article{bai2025ditto,
  title={Scaling Instruction-Based Video Editing with a High-Quality Synthetic Dataset},
  author={Bai, Qingyan and Wang, Qiuyu and Ouyang, Hao and Yu, Yue and Wang, Hanlin and Wang, Wen and Cheng, Ka Leong and Ma, Shuailei and Zeng, Yanhong and Liu, Zichen and Xu, Yinghao and Shen, Yujun and Chen, Qifeng},
  journal={arXiv preprint arXiv:2510.15742},
  year={2025}
}

@article{tan2025omnivideo,
  title={Omni-video: Democratizing unified video understanding and generation},
  author={Tan, Zhiyu and Yang, Hao and Qin, Luozheng and Gong, Jia and Yang, Mengping and Li, Hao},
  journal={arXiv preprint arXiv:2507.06119},
  year={2025}
}

@article{wei2025univideo,
  title={Univideo: Unified understanding, generation, and editing for videos},
  author={Wei, Cong and Liu, Quande and Ye, Zixuan and Wang, Qiulin and Wang, Xintao and Wan, Pengfei and Gai, Kun and Chen, Wenhu},
  journal={arXiv preprint arXiv:2510.08377},
  year={2025}
}

@article{he2025openve,
  title={OpenVE-3M: A Large-Scale High-Quality Dataset for Instruction-Guided Video Editing},
  author={He, Haoyang and Wang, Jie and Zhang, Jiangning and Xue, Zhucun and Bu, Xingyuan and Yang, Qiangpeng and Wen, Shilei and Xie, Lei},
  journal={arXiv preprint arXiv:2512.07826},
  year={2025}
}

@article{couairon2023videdit,
  title={Videdit: Zero-shot and spatially aware text-driven video editing},
  author={Couairon, Paul and Rambour, Cl{\'e}ment and Haugeard, Jean-Emmanuel and Thome, Nicolas},
  journal={Transactions on Machine Learning Research},
  year={2023}
}

@article{zheng2024open,
  title={Open-sora: Democratizing efficient video production for all},
  author={Zheng, Zangwei and Peng, Xiangyu and Yang, Tianji and Shen, Chenhui and Li, Shenggui and Liu, Hongxin and Zhou, Yukun and Li, Tianyi and You, Yang},
  journal={arXiv preprint arXiv:2412.20404},
  year={2024}
}

@article{wan2025wan,
  title={Wan: Open and advanced large-scale video generative models},
  author={Wan, Team and Wang, Ang and Ai, Baole and Wen, Bin and Mao, Chaojie and Xie, Chen-Wei and Chen, Di and Yu, Feiwu and Zhao, Haiming and Yang, Jianxiao and others},
  journal={arXiv preprint arXiv:2503.20314},
  year={2025}
}

@article{kong2024hunyuanvideo,
  title={Hunyuanvideo: A systematic framework for large video generative models},
  author={Kong, Weijie and Tian, Qi and Zhang, Zijian and Min, Rox and Dai, Zuozhuo and Zhou, Jin and Xiong, Jiangfeng and Li, Xin and Wu, Bo and Zhang, Jianwei and others},
  journal={arXiv preprint arXiv:2412.03603},
  year={2024}
}

@inproceedings{dong2024icl-survey,
  title={A survey on in-context learning},
  author={Dong, Qingxiu and Li, Lei and Dai, Damai and Zheng, Ce and Ma, Jingyuan and Li, Rui and Xia, Heming and Xu, Jingjing and Wu, Zhiyong and Chang, Baobao and others},
  booktitle={Proceedings of the 2024 conference on empirical methods in natural language processing},
  pages={1107--1128},
  year={2024}
}

@inproceedings{zisenorita,
  title={Se{\~n}orita-2M: A High-Quality Instruction-based Dataset for General Video Editing by Video Specialists},
  author={Zi, Bojia and Ruan, Penghui and Chen, Marco and Qi, Xianbiao and Hao, Shaozhe and Zhao, Shihao and Huang, Youze and Liang, Bin and Xiao, Rong and Wong, Kam-Fai},
  booktitle={The Thirty-ninth Annual Conference on Neural Information Processing Systems Datasets and Benchmarks Track},
  year={2025}
}

@inproceedings{huang2024vbench,
  title={Vbench: Comprehensive benchmark suite for video generative models},
  author={Huang, Ziqi and He, Yinan and Yu, Jiashuo and Zhang, Fan and Si, Chenyang and Jiang, Yuming and Zhang, Yuanhan and Wu, Tianxing and Jin, Qingyang and Chanpaisit, Nattapol and others},
  booktitle={Proceedings of the IEEE/CVF Conference on Computer Vision and Pattern Recognition},
  pages={21807--21818},
  year={2024}
}

@article{zheng2025vbench,
  title={Vbench-2.0: Advancing video generation benchmark suite for intrinsic faithfulness},
  author={Zheng, Dian and Huang, Ziqi and Liu, Hongbo and Zou, Kai and He, Yinan and Zhang, Fan and Gu, Lulu and Zhang, Yuanhan and He, Jingwen and Zheng, Wei-Shi and others},
  journal={arXiv preprint arXiv:2503.21755},
  year={2025}
}

@article{schuhmann2022laion,
  title={Laion-5b: An open large-scale dataset for training next generation image-text models},
  author={Schuhmann, Christoph and Beaumont, Romain and Vencu, Richard and Gordon, Cade and Wightman, Ross and Cherti, Mehdi and Coombes, Theo and Katta, Aarush and Mullis, Clayton and Wortsman, Mitchell and others},
  journal={Advances in neural information processing systems},
  volume={35},
  pages={25278--25294},
  year={2022}
}

@article{yang2025qwen3,
  title={Qwen3 technical report},
  author={Yang, An and Li, Anfeng and Yang, Baosong and Zhang, Beichen and Hui, Binyuan and Zheng, Bo and Yu, Bowen and Gao, Chang and Huang, Chengen and Lv, Chenxu and others},
  journal={arXiv preprint arXiv:2505.09388},
  year={2025}
}

@article{bai2025qwen3-vl,
  title={Qwen3-vl technical report},
  author={Bai, Shuai and Cai, Yuxuan and Chen, Ruizhe and Chen, Keqin and Chen, Xionghui and Cheng, Zesen and Deng, Lianghao and Ding, Wei and Gao, Chang and Ge, Chunjiang and others},
  journal={arXiv preprint arXiv:2511.21631},
  year={2025}
}

@misc{flux-2-2025,
    author={Black Forest Labs},
    title={{FLUX.2: Frontier Visual Intelligence}},
    year={2025},
    howpublished={\url{https://bfl.ai/blog/flux-2}},
}

@misc{wu2025qwenimagetechnicalreport,
      title={Qwen-Image Technical Report}, 
      author={Chenfei Wu and Jiahao Li and Jingren Zhou and Junyang Lin and Kaiyuan Gao and Kun Yan and Sheng-ming Yin and Shuai Bai and Xiao Xu and Yilei Chen and Yuxiang Chen and Zecheng Tang and Zekai Zhang and Zhengyi Wang and An Yang and Bowen Yu and Chen Cheng and Dayiheng Liu and Deqing Li and Hang Zhang and Hao Meng and Hu Wei and Jingyuan Ni and Kai Chen and Kuan Cao and Liang Peng and Lin Qu and Minggang Wu and Peng Wang and Shuting Yu and Tingkun Wen and Wensen Feng and Xiaoxiao Xu and Yi Wang and Yichang Zhang and Yongqiang Zhu and Yujia Wu and Yuxuan Cai and Zenan Liu},
      year={2025},
      eprint={2508.02324},
      archivePrefix={arXiv},
      primaryClass={cs.CV},
      url={https://arxiv.org/abs/2508.02324}, 
}

@article{chen2025ivebench,
  title={Ivebench: Modern benchmark suite for instruction-guided video editing assessment},
  author={Chen, Yinan and Zhang, Jiangning and Hu, Teng and Zeng, Yuxiang and Xue, Zhucun and He, Qingdong and Wang, Chengjie and Liu, Yong and Hu, Xiaobin and Yan, Shuicheng},
  journal={arXiv preprint arXiv:2510.11647},
  year={2025}
}

@article{comanici2025gemini,
  title={Gemini 2.5: Pushing the frontier with advanced reasoning, multimodality, long context, and next generation agentic capabilities},
  author={Comanici, Gheorghe and Bieber, Eric and Schaekermann, Mike and Pasupat, Ice and Sachdeva, Noveen and Dhillon, Inderjit and Blistein, Marcel and Ram, Ori and Zhang, Dan and Rosen, Evan and others},
  journal={arXiv preprint arXiv:2507.06261},
  year={2025}
}

@misc{team2025lucy,
  title={Lucy edit: Open-weight text-guided video editing},
  author={Team, DecartAI},
  year={2025},
  publisher={Accessed}
}

@inproceedings{wu2025insvie,
  title={Insvie-1m: Effective instruction-based video editing with elaborate dataset construction},
  author={Wu, Yuhui and Chen, Liyi and Li, Ruibin and Wang, Shihao and Xie, Chenxi and Zhang, Lei},
  booktitle={Proceedings of the IEEE/CVF International Conference on Computer Vision},
  pages={16692--16701},
  year={2025}
}

@inproceedings{jiang2025vace,
  title={Vace: All-in-one video creation and editing},
  author={Jiang, Zeyinzi and Han, Zhen and Mao, Chaojie and Zhang, Jingfeng and Pan, Yulin and Liu, Yu},
  booktitle={Proceedings of the IEEE/CVF International Conference on Computer Vision},
  pages={17191--17202},
  year={2025}
}

@article{von2007tutorial,
  title={A tutorial on spectral clustering},
  author={Von Luxburg, Ulrike},
  journal={Statistics and computing},
  volume={17},
  number={4},
  pages={395--416},
  year={2007},
  publisher={Springer}
}

@inproceedings{ke2021musiq,
  title={Musiq: Multi-scale image quality transformer},
  author={Ke, Junjie and Wang, Qifei and Wang, Yilin and Milanfar, Peyman and Yang, Feng},
  booktitle={Proceedings of the IEEE/CVF international conference on computer vision},
  pages={5148--5157},
  year={2021}
}

@inproceedings{li2023amt,
  title={Amt: All-pairs multi-field transforms for efficient frame interpolation},
  author={Li, Zhen and Zhu, Zuo-Liang and Han, Ling-Hao and Hou, Qibin and Guo, Chun-Le and Cheng, Ming-Ming},
  booktitle={Proceedings of the IEEE/CVF Conference on Computer Vision and Pattern Recognition},
  pages={9801--9810},
  year={2023}
}

@inproceedings{radford2021learning,
  title={Learning transferable visual models from natural language supervision},
  author={Radford, Alec and Kim, Jong Wook and Hallacy, Chris and Ramesh, Aditya and Goh, Gabriel and Agarwal, Sandhini and Sastry, Girish and Askell, Amanda and Mishkin, Pamela and Clark, Jack and others},
  booktitle={International conference on machine learning},
  pages={8748--8763},
  year={2021},
  organization={PmLR}
}

@article{hu2026evolution,
  title={Evolution of Video Generative Foundations},
  author={Hu, Teng and Zhang, Jiangning and Huang, Hongrui and Yi, Ran and Su, Zihan and Weng, Jieyu and Xue, Zhucun and Ma, Lizhuang and Yang, Ming-Hsuan and Tao, Dacheng},
  journal={arXiv preprint arXiv:2604.06339},
  year={2026}
}

@article{hu2025hunyuancustom,
  title={Hunyuancustom: A multimodal-driven architecture for customized video generation},
  author={Hu, Teng and Yu, Zhentao and Zhou, Zhengguang and Liang, Sen and Zhou, Yuan and Lin, Qin and Lu, Qinglin},
  journal={arXiv preprint arXiv:2505.04512},
  year={2025}
}

@article{chen2026cinedance,
  title={CineDance: Towards Next-Generation Multi-Shot Long-Form Cinematic Audio-Video Generation},
  author={Chen, Yuheng and Hu, Teng and Wang, Yuji and He, Qingdong and Xue, Zhucun and Zhou, Qianyu and Li, Jason and Ma, Lizhuang and Zhang, Jiangning and Tao, Dacheng},
  journal={arXiv preprint arXiv:2606.09639},
  year={2026}
}

@article{chen2026omni,
  title={Omni-Customizer: End-to-End MultiModal Customization for Joint Audio-Video Generation},
  author={Chen, Yuheng and He, Qingdong and Hu, Teng and Wang, Yuji and Wang, Yabiao and Ma, Lizhuang and Zhang, Jiangning},
  journal={arXiv preprint arXiv:2605.17488},
  year={2026}
}

@article{wang2026personashot,
  title={PersonaShot: Benchmarking Person-Centric Narrative Continuity in Multi-Shot Video Generation},
  author={Wang, Yuji and Chen, Yuheng and Hu, Teng and Yi, Ran and Hong, Yijia and Feng, Han and Cao, Weijian and Wang, Chengjie and Ma, Lizhuang and Zhang, Jiangning},
  journal={arXiv preprint arXiv:2608.16717},
  year={2026}
}
\bibliographystyle{unsrt}

\appendix
\section*{Appendix Overview}

This appendix provides additional details and results supporting the main paper.

$\bullet$ Appendix~\ref{suppl:related_work} reviews related work on video editing, visual in-context learning, and datasets.

$\bullet$ Appendix~\ref{suppl:dataset} describes the task taxonomy, data curation pipeline, dataset composition, and prompt templates.

$\bullet$ Appendix~\ref{suppl:bench} presents the composition and examples of \benchmark{}.

$\bullet$ Appendix~\ref{suppl:implementation} documents training configurations, baseline input settings, and evaluation protocols, including human evaluation and Reference Adherence.

$\bullet$ Appendix~\ref{suppl:experiments} reports additional quantitative results, controlled analyses, computational efficiency, and qualitative comparisons.

\section{Detailed Related Work}
\label{suppl:related_work}

\subsection{Instruction-driven Video Editing}

\textbf{Training-free methods} adapt pre-trained image diffusion models for video editing without task-specific training. Tune-A-Video~\cite{wu2023tuneavideo} fine-tunes a text-to-video backbone on a single video for personalized editing. TokenFlow~\cite{geyer2023tokenflow} enforces temporal coherence via attention-space nearest-neighbor field matching, while AnyV2V~\cite{ku2024anyv2v} provides a plug-and-play tuning-free framework guided by the first edited frame.

\textbf{End-to-end trained systems} learn editing policies directly from paired datasets to robustly handle complex instructions. EasyV2V~\cite{mai2025easyv2v} highlights training data quality as the primary performance driver across the design space. NOVA~\cite{pan2026nova} decouples editing into sparse control prediction and dense synthesis. VideoCoF~\cite{yang2026videocof} adopts chain-of-thought observe-reason-edit stages for temporal consistency, and ReCo~\cite{zhang2025reco} incorporates bounding-box constraints for localized modifications. However, these methods rely exclusively on text instructions, lacking visual reference integration.

\textbf{Visual reference integration} has recently emerged to bridge this gap. Kiwi-Edit~\cite{lin2026kiwi} introduces a single reference image via fixed learnable queries, which lack content adaptivity. While reference-conditioned or customized video generation architectures~\cite{hu2025hunyuancustom,zhang2026omnitransfer} utilize visual exemplars, they primarily focus on synthesis from noise rather than the direct manipulation of an existing source video. EditVerse~\cite{ju2025editverse} tokenizes heterogeneous tasks through interleaved sequence self-attention, but treats distinct modalities homogeneously. In contrast, \method{} uniquely supports heterogeneous reference inputs (single images, image pairs, and video pairs) via adaptive semantic extraction.

\subsection{In-Context Learning for Visual Generation}

In-context learning (ICL) originated in large language models~\cite{dong2024icl-survey} and has expanded to visual synthesis, steering generation via contextual demonstrations without gradient updates.

\textbf{Image generation.} Painter~\cite{wang2023images} and PromptDiffusion~\cite{wang2023promptdiffusion} formulate visual exemplars as unified sequence prompting. Paint by Example~\cite{yang2023paint}, ImageBrush~\cite{yang2023imagebrush}, and Edit Transfer~\cite{chen2025edittransfer} further adapt ICL to exemplar-guided image editing using source-target demonstration pairs.

\textbf{Video generation.} Extending ICL to videos involves severe temporal consistency and computational challenges~\cite{hu2026evolution}. Early efforts explored prompt-level demonstrations~\cite{fei2024video} or autoregressive zero-shot clip imitation (VICL~\cite{zhang2025vicl}). Recent works branch into several technical paradigms: EditVerse~\cite{ju2025editverse} uses interleaved full self-attention across modalities; TIC-FT~\cite{kim2025temporal} concatenates frames along the temporal axis with learned buffer frames for few-shot adaptation; OmniTransfer~\cite{zhang2026omnitransfer} and UniTransfer~\cite{lei2025unitransfer} decompose reference and target spaces for concept transfer. Related advances also explore customized and long-sequence synthesis across modalities~\cite{chen2026omni,chen2026cinedance}.

Despite these advancements, existing video ICL methods primarily focus on conditional generation from scratch under homogeneous conditions. \method{} instead addresses heterogeneous multimodal reference integration in an instruction-driven video editing setting, combining Modality-Adaptive Semantic Distillation (MASD) with Dual-Context Injection (DCI).

\subsection{Video Editing Datasets}

Video editing benchmarks have evolved alongside model paradigms. Early datasets primarily target text-only instruction editing: VidEdit~\cite{couairon2023videdit} provides web-crawled pairs with spatial masks, and OpenVE~\cite{he2025openve} introduces diverse operations with preference alignments, alongside benchmarks assessing consistency in long-form generation~\cite{wang2026personashot}. For reference-augmented setups, ReCo~\cite{zhang2025reco} annotates spatial boxes for region editing, while Kiwi-Edit~\cite{lin2026kiwi} augments training with single reference images. 

Unlike previous benchmarks constrained by text-only directives or single-modality exemplars, \dataset{} spans ten editing task categories across heterogeneous single-image, image-pair, and video-pair references, totaling approximately 400K raw curated instances before consolidation.
\section{More Details on Data Curation}
\label{suppl:dataset}

\subsection{Task Taxonomy and Definitions}
\begin{figure}[h]
    \centering
    \includegraphics[width=\linewidth]{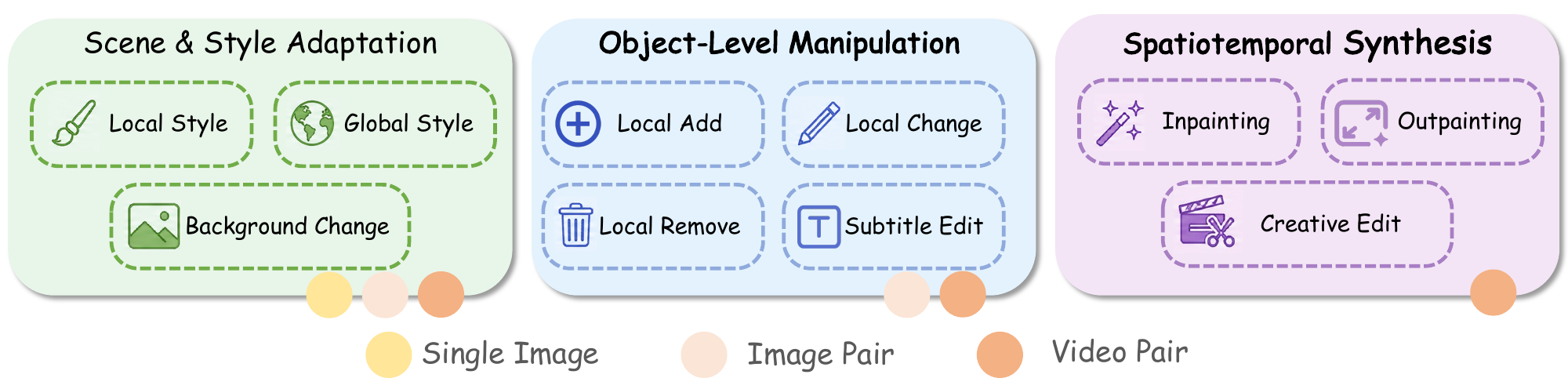}
    \caption{Taxonomy of Editing Tasks and Modalities.}
    \label{fig:data_fig}
\end{figure}
We categorize \dataset{} into three task groups aligned with increasing editing complexity along visual, spatial, and temporal dimensions (see Fig. \ref{fig:data_fig}). This taxonomy determines which reference modalities are supported for each task, ensuring that the model receives the minimal sufficient visual context for the target editing granularity.

\textbf{Scene \& Style Adaptation.} 
This category focuses on global or regional visual transformation, including tasks such as \textit{Global Style}, \textit{Local Style}, and \textit{Background Change}. All three reference modalities are supported. \textbf{Single Image} provides sufficient color, texture, and material priors to serve as an effective anchor for style transfer. \textbf{Image Pair} and \textbf{Video Pair} further clarify transformation boundaries via explicit before/after comparisons, which is particularly beneficial for background replacement or complex lighting transfer where global consistency must be preserved.

\textbf{Object-Level Manipulation.} 
This category targets localized object addition, removal, or attribute modification, covering tasks such as \textit{Local Add}, \textit{Local Change}, \textit{Local Remove}, and \textit{Subtitle Edit}. Only Image Pair and Video Pair are supported. Accurate spatial grounding requires source-target correspondence, which Single Image cannot provide because it lacks the ``before'' context to disambiguate between background preservation and object introduction. \textbf{Image Pair} explicitly defines the spatial displacement and local deformation through source-target mapping. \textbf{Video Pair} further supplies motion priors (e.g., deformation during object motion), which are critical for handling occlusion and temporal interaction in dynamic scenes.

\textbf{Spatiotemporal Synthesis.} 
This category covers structurally complex tasks including \textit{Inpainting}, \textit{Outpainting}, and \textit{Creative Edit}. Only \textbf{Video Pair} is supported. These tasks elevate editing from pixel-level modification to spatiotemporal logic reconstruction. Static references (Single Image or Image Pair) cannot convey action continuity or object evolution. Video Pair provides dense temporal semantic correspondences that constrain the generation process, enabling the model to learn temporal regularities from reference videos and produce physically plausible completion and scene extrapolation.

Table~\ref{tab:task_modality} summarizes the supported modalities for each task type.

\begin{table}[ht]
\centering
\caption{Supported reference modalities for each task type in \dataset{}. SI = Single Image; IP = Image Pair; VP = Video Pair.}
\label{tab:task_modality}
\begin{tabular}{llccc}
\toprule
\textbf{Category} & \textbf{Task Type} & \textbf{SI} & \textbf{IP} & \textbf{VP} \\
\midrule
\multirow{3}{*}{\makecell[l]{Scene \& Style \\ Adaptation}} 
& Global Style & \checkmark & \checkmark & \checkmark \\
& Local Style  & \checkmark & \checkmark & \checkmark \\
& Background Change    & \checkmark & \checkmark & \checkmark \\
\midrule
\multirow{4}{*}{\makecell[l]{Object-Level \\ Manipulation}} 
& Local Add            & $-$       & \checkmark & \checkmark \\
& Local Change         & $-$       & \checkmark & \checkmark \\
& Local Remove         & $-$       & \checkmark & \checkmark \\
& Subtitle Edit        & $-$       & \checkmark & \checkmark \\
\midrule
\multirow{3}{*}{\makecell[l]{Spatiotemporal \\ Synthesis}} 
& Inpainting           & $-$       & $-$       & \checkmark \\
& Outpainting          & $-$       & $-$       & \checkmark \\
& Creative Edit        & $-$       & $-$       & \checkmark \\
\bottomrule
\end{tabular}
\end{table}

\subsection{Dataset Comparison}
\label{suppl:dataset_comparison}

Table~\ref{tab:dataset_comparison} compares \dataset{} with existing video editing datasets. Prior datasets are either text-only or support only a single visual reference modality, limiting their coverage of diverse editing scenarios. In contrast, \dataset{} is the first large-scale dataset that supports three reference modalities (Single Image, Image Pair, and Video Pair) across 10 distinct editing task types, providing a systematic foundation for visual in-context video editing research.

\begin{table}[t] 
\centering 
\caption{Comparison of video editing datasets. \dataset{} provides the most comprehensive support for multi-modal visual in-context references, including Single Image (SI), Image Pair (IP), and Video Pair (VP).} 
\label{tab:dataset_comparison} 
\resizebox{0.9\textwidth}{!}{%
\begin{tabular}{l|c|cccc|c|c|c} 
\toprule 
\multirow{2}{*}{\textbf{Dataset}} & \multirow{2}{*}{\textbf{Scale}} & \multicolumn{4}{c|}{\textbf{Reference Modality}} & \multirow{2}{*}{\begin{tabular}[c]{@{}c@{}}\textbf{Task}\\ \textbf{Types}\end{tabular}} & \multirow{2}{*}{\textbf{Frames}} & \multirow{2}{*}{\textbf{Resolution}} \\
\cline{3-6}
 &  & \textbf{Text} & \textbf{SI} & \textbf{IP} & \textbf{VP} &  &  & \\ 
\midrule 
InsViE~\cite{wu2025insvie} & 1.0M & \checkmark & -- & -- & -- & - & 25 & 1024$\times$576 \\ 
Ditto~\cite{bai2025ditto} & 1.0M & \checkmark & -- & -- & -- & 3 & 101 & 1280$\times$720 \\ 
Se\~norita~\cite{zisenorita} & 2.0M & \checkmark & -- & -- & -- & 9 & 33--64 & 336$\times$592--1120$\times$1984 \\ 
OpenVE~\cite{he2025openve} & 3.0M & \checkmark & -- & -- & -- & 8 & 65--129 & 1280$\times$720 \\ 
ReCo~\cite{zhang2025reco} & 40K & \checkmark & -- & -- & -- & 5 & 81 & 480$\times$832 \\  
RefVIE~\cite{lin2026kiwi} & 477K & \checkmark & \checkmark & -- & -- & 2 & 65--129 & 1280$\times$720 \\
\midrule
\rowcolor{cyan!15} \textbf{\dataset{} (Ours)} & \textbf{400K} & \checkmark & \checkmark & \checkmark & \checkmark & \textbf{10} & \textbf{33--129} & \textbf{480$\times$832--1120$\times$1984} \\ 
\bottomrule 
\end{tabular}
}
\end{table}

\subsection{More Details on Data Pipeline}

The construction of \dataset{} follows a rigorous four-stage pipeline designed to ensure both large-scale diversity and fine-grained semantic alignment.

\textbf{Source Data Filtering.}
We begin by curating a high-quality pool of approximately 1.0M video segments, applying an aesthetic cut-off of 0.5 to filter out low-quality raw footage. Given the scale of the source data, we employ the \textbf{Qwen3-VL-8B} model as an efficient initial rater to assess semantic consistency (see \ref{prompt_template_0}). The filtered corpus is then split: a primary subset serves as the foundation for base editing tasks, while the remainder is reserved for synthesizing heterogeneous in-context exemplars.

\textbf{Instruction Synthesis.}
To derive precise editing instructions, we utilize the more powerful \textbf{Qwen3-VL-32B} model. As detailed in \ref{prompt_template_1} and \ref{prompt_template_2}, our synthesis strategy prioritizes two core objectives: (i) \textit{Semantic Adherence}, ensuring reference content strictly aligns with the editing intent, and (ii) \textit{Visual Divergence}, enforcing a clear distinction between references and source videos to robustly evaluate the model's structural generalization rather than simple memorization.

\textbf{Visual Reference Generation.}
For \textit{Single Image} and \textit{Image Pair} modalities, we feed the synthesized prompts into \textbf{FLUX.2}~\cite{flux-2-2025}, which achieves high-fidelity generation in approximately 4s per GPU, facilitating large-scale production. For \textit{Video Pair} synthesis, we employ a hybrid matching strategy. For common tasks, we use \textit{Identity-based Grouping} followed by intra-group random sampling. For specialized tasks like \textit{Creative Edit}, we perform semantic clustering using \textbf{BGE-large} embeddings and \textbf{HDBSCAN}. This allows us to construct a \textbf{Semantic Knowledge Graph} from which relevant in-context pairs are strategically sampled.

\textbf{Semantic Verification.}
In the final stage, we perform a multi-dimensional audit covering \textit{Instruction Compliance}, \textit{Detail Fidelity}, and \textit{Visual Quality}. To ensure maximum judgment reliability for the final dataset, we upgrade the evaluator to \textbf{Qwen3-VL-32B}, retaining only samples that exhibit superior average scores across all dimensions. This closed-loop verification guarantees that every triplet in \dataset{} provides a meaningful and accurate signal for in-context learning.

\subsection{Prompt Template in Data Pipeline}
\label{suppl:prompt_template}

To support diverse in-context editing granularities, our data construction pipeline employs a structured, modality-aware prompting strategy using Qwen3-VL. We design distinct templates tailored to the specific requirements of heterogeneous in-context modalities, with a particular emphasis on ensuring that synthesized references remain \textit{thematically consistent yet visually distinct} from the source video content to enhance model generalization.

\subsubsection{VLM-as-a-Judge: Multi-Dimensional Quality Assessment}
\label{prompt_template_0}

To ensure the high-quality synthesis of \dataset{}, we employ a VLM-as-a-Judge mechanism. Unlike simple heuristic-based filtering, our approach leverages the advanced reasoning capabilities of Qwen3-VL to perform a holistic evaluation of the triplet: (Source Video, Edited Video, Instruction). This stage acts as a critical quality gate in our data curation pipeline, filtering out low-fidelity generations and ensuring precise instruction-image-video alignment.

The assessment logic is grounded in three orthogonal dimensions: Instruction Compliance (semantic accuracy), Consistency \& Detail Fidelity (spatial-temporal identity preservation), and Visual Quality \& Stability (generative aesthetics). We specifically enforce a "bottleneck constraint" where structural or aesthetic scores cannot exceed the compliance score, preventing the inclusion of visually pleasing but semantically irrelevant samples.
\begin{promptbox}{Video Score Template}
You are a data rater specializing in grading video editing quality. You will be given two videos (before and after editing) and the editing instruction. Your task is to evaluate the video edit on a 5-point scale from three perspectives:

\textbf{Instruction Compliance} \\
1. No edit or completely unrelated change.\\
2. Partial or wrong edit; instruction largely unfollowed.\\
3. Instruction mostly followed but with significant errors or omissions.\\
4. Correct edit with only minor inaccuracies.\\
5. Perfect edit that fully matches the instruction.

\textbf{Consistency \& Detail Fidelity} \\
1. Major content lost, distorted, or hallucinated; original scene barely recognisable. \\
2. Main subject recognisable but key details wrong/missing. \\
3. Overall structure correct; some local artifacts, warping, or minor omissions. \\
4. Nearly all geometry and motion intact; only slight, non-distracting deformation. \\
5. All objects, spatial relations, and motion perfectly preserved outside the edit region.

\textbf{Visual Quality \& Stability} \\
1. Severe flickering, artifacts, or temporal inconsistency making the video unwatchable.\\
2. Significant and distracting quality degradation or temporal instability.\\
3. Noticeable but tolerable flaws or flickering.\\
4. Largely stable and high quality with only minor, subtle issues.\\
5. Perfectly stable and temporally coherent; indistinguishable from real footage. \\

Note: The scores for Consistency \& Detail Fidelity and Visual Quality \& Stability should not be higher than the Instruction Compliance score.

Example Response Format
Brief reasoning: A short explanation of the scores, no more than 30 words.
Instruction Compliance: A number from 1 to 5.
Consistency \& Detail Fidelity: A number from 1 to 5.
Visual Quality \& Stability: A number from 1 to 5.
Editing instruction is: \{edit\_prompt\}.

Below are the videos before and after editing:
\end{promptbox}

\subsubsection{Cross-Scenario Reference Synthesis for Single-Image Modality}
\label{prompt_template_1}
This part of the pipeline focuses on generating a standalone \textbf{Visual Exemplar} ($R$) to support the single-image reference modality. To prevent the model from overfitting to specific instances, we adopt a \textbf{``Category-Consistent but Content-Divergent''} strategy. The MLLM is strictly instructed to avoid reproducing the exact scene from the original video, instead generating objects or environments that share the same underlying concept but differ in specific visual instantiation.

\begin{promptbox}{Template for Single-Image Reference Synthesis}
\textbf{Core Constraint:} The output must describe a \textit{different} but \textit{same-type} instance compared to the instruction.

\begin{itemize}
    \item \textbf{Object-Level Manipulation:} Generate a detailed prompt for a DIFFERENT object of the same category (e.g., if the instruction adds a ``brown fedora,'' describe a ``red beret''). Focus strictly on the object’s material, shape, and texture while ensuring no background context is included.
    \item \textbf{Style Adaptation:} Use a ``Style-on-Random-Scene'' approach. Pick a RANDOM everyday scene unrelated to the original video (e.g., ``a cat on a windowsill'') and render it in the target style to isolate stylistic characteristics from content.
    \item \textbf{Scene Adaptation:} Describe a THEMATICALLY SIMILAR but visually distinct background (e.g., a ``retro cyberpunk alley'' to represent a ``neon futuristic city'' theme), ensuring all foreground subjects are removed.
\end{itemize}
\end{promptbox}

\subsubsection{Visual Task Distillation for Image-Pair Modality}
\label{prompt_template_2}
For the image-pair modality, the objective is to transform dynamic temporal editing logic into static, paired visual representations. By processing the original frame, the edited frame, and the instruction simultaneously, the MLLM distills the dynamic transition into a dense semantic mapping that captures the ``before-and-after'' logic.

\begin{promptbox}{Template for Image-Pair Task Distillation}
\textbf{System Prompt:} \\
``You are an expert at converting video editing tasks into equivalent image editing tasks. Analyze the visual difference between the ORIGINAL and EDITED frames, understand the editing logic, and generate a representative image editing task.''

\medskip
\textbf{User Inputs:}
\begin{itemize}
    \item \textit{Visual Anchor:} ORIGINAL video frame (pre-edit).
    \item \textit{Target Anchor:} EDITED video frame (post-edit).
    \item \textit{Editing Logic:} Original video instruction: ``$P$''.
\end{itemize}

\medskip
\textbf{Synthesized Outputs:} 
\begin{itemize}
    \item \textbf{IMAGE\_PROMPT}: A detailed description of the source scene tailored for high-fidelity T2I generation.
    \item \textbf{EDIT\_INSTRUCTION}: A clear, distilled instruction describing the spatial or stylistic transformation required to bridge the two frames.
\end{itemize}
\end{promptbox}

Through these targeted prompting strategies, \dataset{} provides a systematic foundation for training models to interpret and execute edits across diverse visual contexts without relying on simple memorization.
\section{Comprehensive Specifications of \benchmark{}}
\label{suppl:bench}
\begin{figure}[ht]
    \centering
    \includegraphics[width=\linewidth]{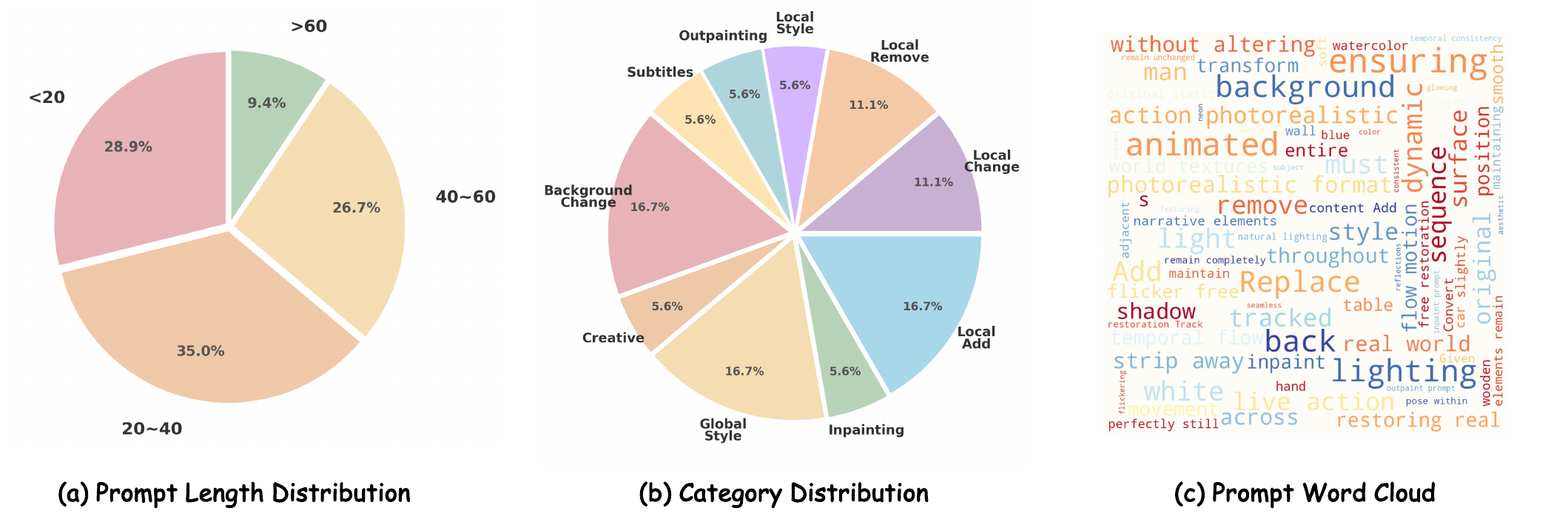}
    \caption{Statistics of \benchmark{}.}
    \label{fig:stat_bench}
\end{figure}
As illustrated in Figure~\ref{fig:stat_bench}, the design of \benchmark{} is strategically aligned with the requirements of visual in-context editing. 
The category distribution (Fig.~\ref{fig:stat_bench}b) spans a diverse spectrum from global stylistic adaptations to fine-grained local manipulations, necessitating the synergistic use of both textual instructions and visual exemplars. 
Furthermore, the prevalence of detailed, long-context prompts (Fig.~\ref{fig:stat_bench}a) and the rich semantic vocabulary (Fig.~\ref{fig:stat_bench}c) ensure a rigorous evaluation of the model's ability to resolve linguistic ambiguity through visual anchoring. 
Overall, \benchmark{} provides a comprehensive and challenging testbed that effectively validates the efficacy of our visual in-context editing paradigm.

\begin{figure}[ht]
    \centering
    \includegraphics[width=\linewidth]{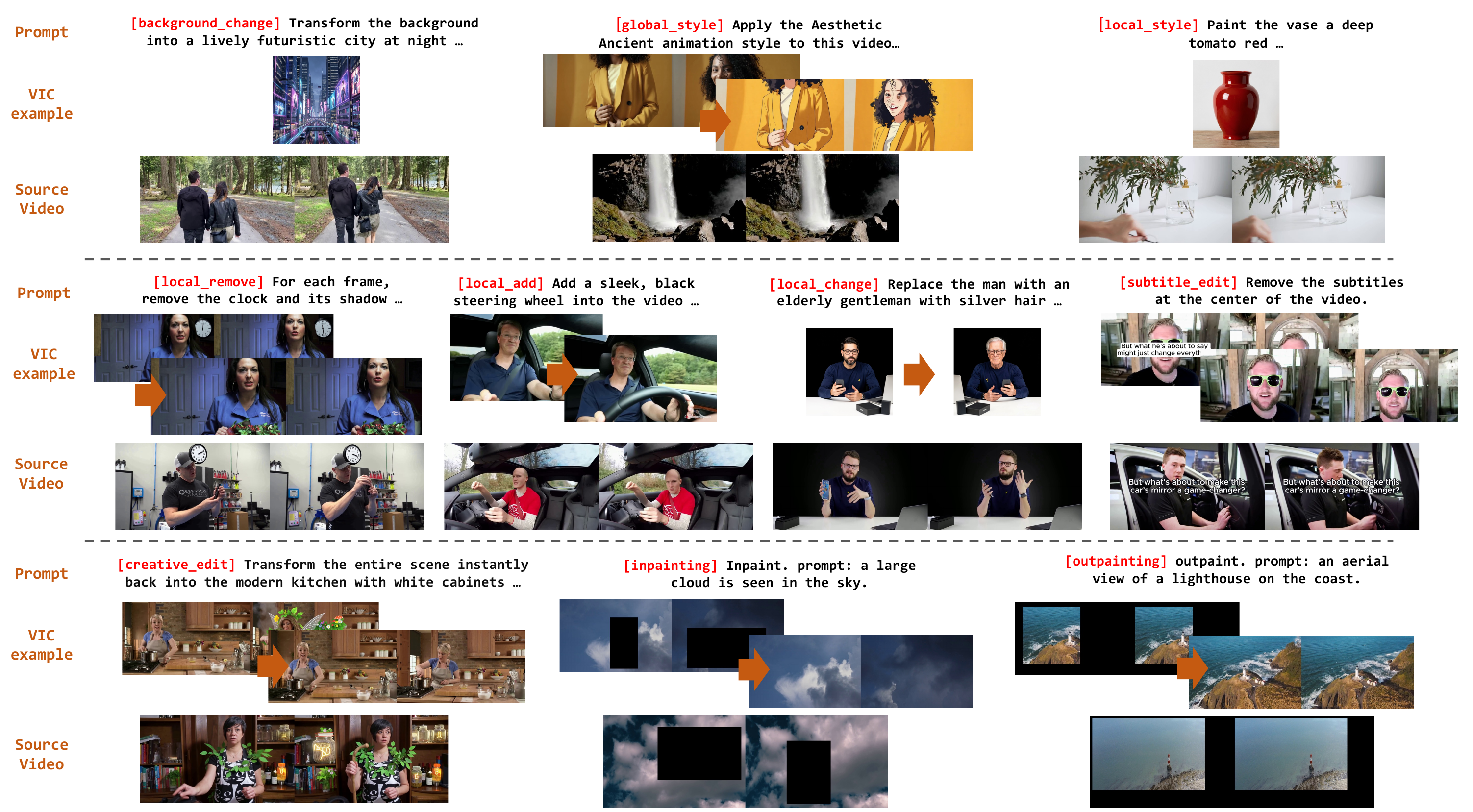}
    \caption{Examples of \benchmark{}.}
    \label{fig:bench_example}
\end{figure}

Fig. \ref{fig:bench_example} showcases the diverse range of video editing tasks covered by VicEdit-Bench.

\section{Implementation and Evaluation Details}
\label{suppl:implementation}
\subsection{Detailed Training Configurations}
\label{suppl:training}

Following the three-stage training strategy described in Section \ref{sec:methodology}, we progressively incorporate in-context editing capability into VicEdit. The visual-reference and instruction branches both use pretrained Qwen weights as their base; the language components are adapted with LoRA in Stages 2 and 3, while Stage 1 keeps the MLLM frozen.

\textbf{Stage 1: Base Editing.} We fully fine-tune the Video DiT on text-instructed editing tasks to establish fundamental instruction-following ability. The model is conditioned on the source video and text instruction,
with the target video providing training supervision; no visual reference is used in this stage. Only the DiT parameters are updated, while the MLLM remains frozen
and no MLLM LoRA adapters are trained. We train for 2 epochs with a learning rate of $1\times10^{-5}$.
This run uses the 70,000-record instruction subset, comprising 30,000 OpenVE, 30,000 Se\~norita, and 10,000 OpenVE-derived Creative Edit examples. It is distinct from the balanced 200,000-record curation inventory. Each training sample contains 81 frames.

\textbf{Stage 2: In-Context Branch.} We freeze the Video DiT and train the newly introduced context modules, including MASD and DCI. The ICL manifest contains 200,070 records: 60,000 SI, 100,000 IP, and 40,070 VP examples.
We also adapt the language component of the MLLM using rank-64 LoRA adapters, while keeping its underlying pretrained weights
and vision encoder frozen. Each training batch mixes samples from the Single Image, Image Pair, and Video Pair modalities to train the
modality-specific embeddings jointly. We train the context modules and MLLM LoRA adapters for 4 epochs with a learning rate of $1\times10^{-5}$.

\textbf{Stage 3: Joint Fine-Tuning.} We jointly optimize the context modules, rank-64 LoRA adapters in the MLLM language component, and rank-32 LoRA adapters in the Video DiT. The underlying backbone weights of both models and the MLLM vision encoder remain frozen during this stage. We train for 4 epochs with a learning rate of $1\times10^{-5}$.

All stages use 16 GPUs, DeepSpeed ZeRO-3, and BF16 mixed precision, with a reported total training cost of approximately 32 GPU-days.

\subsection{Baselines}
\label{suppl:baselines}

We compare \method{} against six state-of-the-art video editing methods.
For fair comparison, all methods are evaluated at their recommended resolution
and frame length following their official configurations.

\textbf{VACE~\cite{jiang2025vace}} is a unified video creation and editing
framework based on a 14B Wan2.1 DiT backbone. It formulates diverse editing
tasks as sequence completion problems via a task-agnostic Video Condition Unit
(VCU), supporting both text-only and reference-guided editing at 720p
resolution. However, its single cross-attention mechanism lacks
modality-adaptive semantic extraction for heterogeneous references.

\textbf{Lucy-Edit~\cite{team2025lucy}} is an open-weight text-guided video
editing model built on the Wan2.2-5B backbone. It achieves strong
instruction-following fidelity via attention-based spatial control, but
supports only text-only instructions without visual reference conditioning.

\textbf{DITTO~\cite{bai2025ditto}} builds upon a VACE-based diffusion
transformer (Wan2.2) fine-tuned on the large-scale Ditto-1M dataset ($\sim$1M
paired triplets). Training uses AdamW at $1\times10^{-4}$ for 16K steps on
64 GPUs, processing 101-frame videos at 1280$\times$720 resolution. It
operates exclusively in a text-only paradigm.

\textbf{NovaEdit~\cite{pan2026nova}} (CVPR 2026) proposes a sparse-control,
dense-synthesis strategy built on the 1.3B Wan2.1 VACE backbone, with
cross-attention modules connecting a sparse VACE branch and a dense DiT
branch. Training uses only 5K curated video clips at 832$\times$480
resolution, 81 frames, for $\sim$8K AdamW steps. Visual reference guidance
is not supported.

\textbf{VideoCoF~\cite{yang2026videocof}} formulates video editing as a
temporal chain-of-frames reasoning problem on the 14B Wan-T2V backbone. It
trains on only 50K video pairs at 33 frames across multi-bucket resolutions
(336$\times$592 to 400$\times$944) for 8K steps on 16$\times$H100 GPUs,
and generalizes to 500+ frames at inference via RoPE alignment. The method
supports text-only instructions.

\textbf{Kiwi-Edit~\cite{lin2026kiwi}} is the closest baseline, employing a
dual-backbone architecture of Qwen2.5-VL-3B as the MLLM encoder and
Wan2.2-TI2V-5B as the video DiT. It introduces learnable query tokens (256/512
for instruction, 768 for reference) that extract visual semantics from a single
reference image. Training uses a three-stage curriculum (12K total steps at
$2\times10^{-5}$ LR) up to 1280$\times$704 at 81 frames. Unlike
\method{}, Kiwi-Edit is limited to single-image reference and does not support
image-pair or video-pair modalities.

In contrast, \method{} supports three reference modalities (Single Image,
Image Pair, Video Pair) within a unified framework, and introduces
modality-adaptive semantic distillation (MASD) and dual-context injection
(DCI) to handle heterogeneous visual references.

\paragraph{Reference-to-Text Instruction Enhancement.}
For the main VIC comparison, we use Qwen3-VL~\cite{bai2025qwen3-vl} to enrich the editing instructions supplied to text-only baselines. It receives the original instruction and visual reference, without the source video, and adds editing-relevant visual details while preserving the original editing intent. For paired references, it describes the demonstrated before--after transformation. For illustration, ``convert to Gongbi style'' can be supplemented with attributes such as even coloring and fine brushwork. Text conversion can lose visual information and is not strictly equivalent to providing the raw reference. The enhancement request is summarized below.
\begin{promptbox}{Concise Instruction Enhancement Request}
Refine the editing instruction using the visual example. Add concrete, editing-relevant details while preserving the original intent. For paired examples, describe the demonstrated change. Return only the refined instruction.
\end{promptbox}

\subsection{Evaluation Details}
\label{suppl:evaluation}
We employ five core metrics from the VBench~\cite{huang2024vbench} suite to assess the generative quality and temporal consistency.

\textbf{Background Consistency (BC $\uparrow$)} quantifies the temporal stability of static environments. It computes the average cosine similarity between consecutive frame-level features extracted by a \textbf{CLIP-ViT-B/32}~\cite{radford2021learning} encoder, ensuring the background remains coherent despite camera or subject motion.

\textbf{Motion Smoothness (MS $\uparrow$)} evaluates the physical plausibility of object trajectories. By leveraging motion priors from the \textbf{AMT}~\cite{li2023amt} video frame interpolation model, it detects unnatural jitter or "teleportation" artifacts to ensure fluid pixel-level displacements.

\textbf{Temporal Flickering (TF $\uparrow$)} focuses on high-frequency luminance or texture instabilities. It calculates the pixel-wise standard deviation at identical spatial coordinates across the video duration, penalizing rapid, inconsistent fluctuations often caused by sampling instability.

\textbf{Imaging Quality (IQ $\uparrow$)} assesses low-level physical attributes such as sharpness and noise. It utilizes the \textbf{MUSIQ}~\cite{ke2021musiq} transformer, pre-trained on human-rated data, to objectively measure distortions including motion blur and compression artifacts.

\textbf{Aesthetic Quality (AQ $\uparrow$)} measures the artistic appeal and visual harmony (e.g., composition and lighting). Using a predictor aligned with \textbf{LAION-Aesthetics}~\cite{schuhmann2022laion}, it quantifies how well the generated frames conform to professional photography and cinematic standards.

\paragraph{VLM-Based Editing Scores.}
For the main evaluation, the VLM judge compares sampled frames from the source video and edited output to assess instruction compliance (IC), detail fidelity (DF), and visual quality (VQ). The visual reference is not provided to the main judge. Overall is the arithmetic mean of IC, DF, and VQ. A separate reference-aware evaluation additionally supplies the intended visual reference and identifies its modality and role in specifying the edit; it reports Reference Adherence (RA) separately from Overall.

\paragraph{Human Evaluation Protocol.}
We report two pairwise preference studies, involving ten annotators who each evaluate 20 examples per setting: base editing on OpenVE-Ext and visual in-context editing on \benchmark{}. For each selected task, annotators compare VicEdit outputs with those of the strongest baseline for that task. The selected examples cover all task categories; the VIC examples are approximately balanced across SI, IP, and VP references. Base editing is assessed on instruction compliance, detail fidelity, and visual quality; VIC editing is assessed on instruction compliance, detail fidelity, and reference adherence.

Annotators choose one of three outcomes: VicEdit wins, VicEdit loses, or a tie. A win gives one point to the preferred method; a tie gives each method 0.5 points. VicEdit's tie-adjusted preference score is
\begin{equation}
S_{\mathrm{human}}=100\times\frac{N_{\mathrm{win}}+0.5N_{\mathrm{tie}}}{N_{\mathrm{win}}+N_{\mathrm{lose}}+N_{\mathrm{tie}}},
\label{eq:human_preference}
\end{equation}
where the counts refer to the collected pairwise judgments. Table~\ref{tab:human_eval} reports preference scores of 76\% and 72\% for base editing and VIC editing, respectively. For the same ten-annotator study, human--Gemini agreement is 80\% and 76\%, and Fleiss' $\kappa$ is 0.70 and 0.66, respectively. Human--Gemini agreement measures consistency between human preferences and the automated judge; Fleiss' $\kappa$ measures agreement among annotators on commonly rated items.
\section{Supplementary Experiments}
\label{suppl:experiments}
\subsection{More Metrics of Main Experiments}

\begin{table}[t]
\centering
\caption{Quantitative comparison on \benchmark{} across VLM-based assessments and VBench~\cite{huang2024vbench} dimensions.}
\label{tab:main_metrics}
\small
\setlength{\tabcolsep}{3pt}
\begin{tabular}{c|cccc|ccccc}
\toprule
 & \multicolumn{4}{c|}{\textbf{VLM Evaluation}} & \multicolumn{5}{c}{\textbf{VBench Metrics}} \\
\cmidrule(lr){2-5} \cmidrule(lr){6-10}
Method & Overall$\uparrow$ & \makecell[c]{Instr.\\Comp.$\uparrow$} & \makecell[c]{Detail\\Fid.$\uparrow$} & \makecell[c]{Visual\\Qual.$\uparrow$} & BC$\uparrow$ & MS$\uparrow$ & TF$\uparrow$ & IQ$\uparrow$ & AQ$\uparrow$ \\
\midrule
VACE~\cite{jiang2025vace} & 1.85 & 1.98 & 1.79 & 1.80 & 0.954 & 0.994 & 0.983 & 0.692 & 0.543 \\
Lucy-Edit~\cite{team2025lucy} & 2.25 & 2.40 & 2.20 & 2.15 & 0.962 & 0.993 & 0.983 & 0.701 & 0.544\\
DITTO~\cite{bai2025ditto} & 2.33 & 2.44 & 2.32 & 2.20 & 0.966 & 0.995 & 0.984 & 0.703 & 0.560 \\
NovaEdit~\cite{pan2026nova} & 2.61 & 2.78 & 2.49 &2.54 & 0.969 & 0.994 & 0.985 & 0.707 & 0.562 \\
VideoCoF~\cite{yang2026videocof} & 2.90 & 3.02 & 2.85 & 2.85 & 0.972 & 0.995 & 0.986 & 0.705 & 0.561 \\
Kiwi-Edit & 2.95 & 3.10 & 2.98 & 2.76	& \textbf{0.995} & 0.986	& 0.711 & 0.564 \\
\midrule
\rowcolor{cyan!10}
VicEdit (Ours) & \textbf{3.45} & 3.65 & 3.34 & 3.35 & \textbf{0.975} & 0.994 & \textbf{0.987} & \textbf{0.718} & \textbf{0.568}  \\
\bottomrule
\end{tabular}
\end{table}
\begin{table}[t]
\centering
\caption{Quantitative Comparison on OpenVE-Ext with Qwen-VL-32B~\cite{bai2025qwen3-vl}.}
\label{tab:vlm_eval}
\resizebox{\textwidth}{!}{%
\begin{tabular}{l|c|ccc|cccc|ccc}
\toprule
 & & \multicolumn{3}{c|}{\textbf{Scene \& Style Adaptation}} & \multicolumn{4}{c|}{\textbf{Object-Level Manipulation}} & \multicolumn{3}{c}{\textbf{Spatiotemporal Synthesis}} \\
\cmidrule(lr){3-5} \cmidrule(lr){6-9} \cmidrule(lr){10-12}
\textbf{Methods} 
& \textbf{Overall} & \makecell[c]{\textbf{Global}\\\textbf{Style}} & \makecell[c]{\textbf{Local}\\\textbf{Style}} & \makecell[c]{\textbf{Background}\\\textbf{Change}} 
& \makecell[c]{\textbf{Local}\\\textbf{Change}} & \makecell[c]{\textbf{Local}\\\textbf{Remove}} & \makecell[c]{\textbf{Local}\\\textbf{Add}} 
& \makecell[c]{\textbf{Subtitle}\\\textbf{Edit}} & \makecell[c]{\textbf{Creative}\\\textbf{Edit}} & \textbf{Inpainting} & \textbf{Outpainting} \\
\midrule
VACE~\cite{jiang2025vace} & 2.45 & 2.11 & 1.94 & 2.38 & 2.87 & 2.41 & 2.13 & 2.24 & 2.06 & 2.72 & 3.09 \\
Lucy-Edit~\cite{team2025lucy} & 3.56 & 3.04 & 2.82 & 4.59 & 4.52 & 4.44 & 4.00 & 2.41 & 3.16 & 3.48 & 3.15 \\
DITTO~\cite{bai2025ditto} & 3.83 & \textbf{4.64} & \textbf{3.87} & 4.66 & 4.31 & 4.65 & 3.99 & 3.61 & 1.94 & 2.92 & 3.75 \\
NovaEdit~\cite{pan2026nova} & 3.23 & 1.70 & 2.90 & 4.50 & 3.70 & 4.80 & 1.95 & 3.74 & 2.48 & 3.25 & 3.23 \\
VideoCoF~\cite{yang2026videocof} & 3.41 & 1.75 & 3.20 & 4.55 & 4.60 & 4.45 & 2.90 & 3.82 & 2.31 & 3.15 & 3.35 \\
Kiwi-Edit & 4.14 & 3.90 & 3.10 & 4.80 & 4.95 & 4.80 & 4.45 & 4.12 & 3.98 & 3.75 & 3.58 \\
\midrule 
\rowcolor{cyan!15}
\textbf{VicEdit (Ours)} & \textbf{4.40} & 4.15 & 3.68 & \textbf{4.90} & \textbf{4.98} & \textbf{4.92} & \textbf{4.65} & \textbf{4.38} & \textbf{4.05} & \textbf{4.02} & \textbf{4.23} \\
\bottomrule
\end{tabular}
}
\end{table}

\begin{table}[t]
\centering
\caption{Quantitative comparison on \benchmark{} with Qwen-VL-32B~\cite{bai2025qwen3-vl}.}
\label{tab:our_bench_qwen}
\resizebox{\textwidth}{!}{%
\begin{tabular}{l|c|ccc|cccc|ccc}
\toprule
 & & \multicolumn{3}{c|}{\textbf{Scene \& Style Adaptation}} & \multicolumn{4}{c|}{\textbf{Object-Level Manipulation}} & \multicolumn{3}{c}{\textbf{Spatiotemporal Synthesis}} \\
\cmidrule(lr){3-5} \cmidrule(lr){6-9} \cmidrule(lr){10-12}
\textbf{Methods} 
& \textbf{Overall} & \makecell[c]{\textbf{Global}\\\textbf{Style}} & \makecell[c]{\textbf{Local}\\\textbf{Style}} & \makecell[c]{\textbf{Background}\\\textbf{Change}} 
& \makecell[c]{\textbf{Local}\\\textbf{Change}} & \makecell[c]{\textbf{Local}\\\textbf{Remove}} & \makecell[c]{\textbf{Local}\\\textbf{Add}} 
& \makecell[c]{\textbf{Subtitle}\\\textbf{Edit}} & \makecell[c]{\textbf{Creative}\\\textbf{Edit}} & \textbf{Inpainting} & \textbf{Outpainting} \\
\midrule
VACE~\cite{jiang2025vace} & 2.62 & 2.15 & 2.44 & 2.30 & 3.10 & 2.45 & 2.18 & 2.32 & 2.85 & 3.05 & 3.40 \\
Lucy-Edit~\cite{team2025lucy} & 3.15 & 2.80 & 2.35 & 2.92 & 4.25 & 3.12 & 3.30 & 2.28 & 3.20 & 3.55 & 3.75\\
DITTO~\cite{bai2025ditto} & 3.32 & 3.72 & \textbf{4.21} & 3.35 & 3.25 & 2.68 & 2.35 & 3.42 & 2.10 & 3.30 & 3.12\\
NovaEdit~\cite{pan2026nova} & 3.48 & 2.95 & 3.08 & 3.84 & 3.92 & 4.58 & 3.12 & 3.36 & 2.95 & 3.48 & 3.52 \\
VideoCoF~\cite{yang2026videocof} & 3.81 & 3.52 & 3.31 & 3.75 & 4.45 & 4.38 & 3.61 & \textbf{4.38} & 2.64 & 3.92 & 4.10   \\
Kiwi-Edit~\cite{lin2026kiwi} & 3.96 & 3.98 & 3.28 & 3.82 & \textbf{4.82} & 4.25 & 3.48 & 3.95 & 3.51 & 4.08 & 3.62 \\
\midrule 
\rowcolor{cyan!15}
\textbf{VicEdit (Ours)} & \textbf{4.36} & \textbf{4.25} & 3.85 & \textbf{4.12} & 4.75 & \textbf{4.65} & \textbf{4.32} & 4.30 & \textbf{4.28} & \textbf{4.31} & \textbf{4.42} \\
\bottomrule
\end{tabular}
}
\end{table}
In this section, we supplement more metrics of our main experiments on \benchmark{}. Table \ref{tab:main_metrics} presents the quantitative comparison, demonstrating that our VicEdit outperforms all baseline methods across various VBench dimensions and VLM-based assessments, achieving state-of-the-art results.

\subsection{VLM-based Evaluation Analysis}

To verify the robustness of our evaluation protocol, we introduce Qwen-VL-32B as an alternative grader to compare with Gemini 2.5 Pro.

\textbf{Cross-Model Stability.} As shown in Tables \ref{tab:vlm_eval} and \ref{tab:our_bench_qwen}, the results from Qwen-VL-32B maintain a high degree of correlation with the Gemini-based scores. While Qwen-VL exhibits a higher scoring baseline (typically $+1.0$ to $+1.5$), the relative performance ranking across all methods remains remarkably consistent. This stability suggests that the VLM-based metrics are model-agnostic and provide a reliable reference for video editing quality.

\textbf{Quantitative Results.} Under this cross-model validation, VicEdit (Ours) consistently achieves the highest overall scores (4.40 and 4.36). Our framework demonstrates a clear advantage in complex tasks such as \textit{Local Remove}, \textit{Creative Edit}, and \textit{Outpainting}. While specialized baselines like DITTO and Kiwi-Edit show competitive results in specific dimensions (e.g., \textit{Local Style} and \textit{Local Change}), VicEdit provides the most balanced performance across all editing categories. These findings confirm the efficacy of our unified in-context learning paradigm in producing high-fidelity and temporally consistent video edits.

\subsection{Additional Controlled Analyses from Rebuttal}
\label{suppl:review_controls}
The following tables collect additional modality-wise, prompt-format, and reference-aware analyses. Protocol distinctions and remaining verification items are described below.
\begin{table}[ht]
\centering\small
\caption{Overall scores by reference modality. The SI, IP, and VP subsets contain 30, 70, and 100 examples, respectively, with different task compositions. Comparison with the main Overall score requires the corresponding weights for each task and reference modality.}
\label{tab:modality_breakdown}
\begin{tabular}{lcc}
\toprule Reference modality & Kiwi-Edit & VicEdit \\
\midrule Single Image & 3.02 & 3.56 \\
Image Pair & 2.92 & 3.42 \\
Video Pair & 2.96 & 3.30 \\
\bottomrule
\end{tabular}
\end{table}

\begin{table}[ht]
\centering\small
\caption{Text/visual input comparison reported during rebuttal. The Stage-1 text-only row uses a Kiwi-Edit-style architecture with our Stage-1 checkpoint, matching the architecture and weights of the original-instruction baseline but using enhanced instructions in a separate experiment. The original-instruction result is reported in Table~\ref{tab:visual_reference_effectiveness}. The full-system result is included for context; the comparison is not established as isolating visual access under matched generation and evaluation protocols.}
\label{tab:enhanced_text_control}
\begin{tabular}{lcccc}
\toprule Setting & Overall & Instr. Comp. & Detail Fid. & Visual Qual. \\
\midrule Kiwi-Edit, enhanced text & 2.95 & 3.10 & 2.98 & 2.76 \\
Stage-1 text-only, enhanced text & 3.10 & 3.12 & 3.08 & 3.10 \\
VicEdit, full & 3.45 & 3.65 & 3.34 & 3.35 \\
\bottomrule
\end{tabular}
\end{table}

\begin{table}[ht]
\centering\small
\caption{VideoCoF prompt-format comparison reported during rebuttal. Both variants use reference-derived text and no raw visual exemplar. Verify the actual prompt templates and evaluation records.}
\label{tab:videocof_prompt_control}
\begin{tabular}{lcccc}
\toprule Prompt setting & Overall & Instr. Comp. & Detail Fid. & Visual Qual. \\
\midrule Enriched instruction & 2.90 & 3.02 & 2.85 & 2.85 \\
Temporal Triptych formatting & 3.15 & 3.25 & 3.09 & 3.12 \\
\bottomrule
\end{tabular}
\end{table}

\begin{table}[ht]
\centering\small
\caption{Reference-aware comparison reported during rebuttal, evaluated on the same complete set of 200 samples from \benchmark{} for both methods. The generated videos are reused from the main evaluation without regeneration; the judge inputs and scoring protocol differ. The reference-aware protocol and reconstructed scoring rubric are documented in Appendix~\ref{suppl:evaluation}.}
\label{tab:reference_adherence_comparison}
\begin{tabular}{lcccc}
\toprule Method & Instr. Comp. & Detail Fid. & Visual Qual. & Ref. Adherence \\
\midrule Kiwi-Edit & 3.02 & 2.90 & 2.68 & 2.75 \\
VicEdit & 3.58 & 3.28 & 3.28 & 3.42 \\
\bottomrule
\end{tabular}
\end{table}

\paragraph{Modality-Wise Results.}
Table~\ref{tab:modality_breakdown} reports higher VicEdit scores on SI, IP, and VP. The subsets contain different task types, so their scores do not isolate modality difficulty or the benefit of changing modality on the same task.

\paragraph{Enhanced Text and Visual Access.}
Table~\ref{tab:enhanced_text_control} records the reported text-only and full-model results. A comparison isolates visual access only when source video, instruction text, checkpoint, and generation/evaluation settings are held fixed.
The Stage-1 text-only row (Overall 3.10) is a separate experiment using enhanced instructions at generation time. It uses the same Kiwi-Edit-style architecture and our Stage-1 checkpoint as the original-instruction text-only baseline. It must be distinguished from the original-instruction Stage-1 Kiwi-style text-only baseline in Table~\ref{tab:visual_reference_effectiveness}, even though the reported component scores coincide. The table compares these text-only configurations with the reported full-system result; the score difference alone should not be assigned entirely to visual access without matching the generation and evaluation protocols.

\paragraph{Prompt Formatting.}
Table~\ref{tab:videocof_prompt_control} records VideoCoF's improvement from 2.90 to 3.15 when using Temporal Triptych formatting. This shows that prompt format affects the comparison. It does not establish that all baseline prompt formats have been optimized or that enriched text preserves every visual cue.

\paragraph{Direct Reference Adherence.}
Table~\ref{tab:reference_adherence_comparison} records the reported RA scores of 2.75 for Kiwi-Edit and 3.42 for VicEdit. Both methods are evaluated on the same 200 benchmark samples using their main-evaluation generated videos, with no regeneration. The reference-aware judge receives sampled source and output frames, the enhanced instruction, and the correct reference with its modality identified. The main judge does not receive the visual reference, so these scores are reported separately, as described in Appendix~\ref{suppl:evaluation}.

\subsection{Computational Efficiency}

As shown in Table~\ref{tab:computational_efficiency}, VicEdit achieves inference time comparable to text-only methods with the same backbone. The DiT denoising stage remains unchanged at $\sim$4 min, while the additional MLLM reference feature extraction incurs only 0.5--1.0 s across all three modalities. This overhead represents less than 0.5\% of the total inference time, confirming that both the MASD and DCI modules are computationally lightweight and do not compromise inference efficiency.

\begin{table}[!t]
    \centering
    \caption{Quantitative Comparison of Computational Efficiency. All metrics are measured on a single GPU with a target video length of 81 frames ($480 \times 832$). MLLM Time denotes the reference feature extraction time; DiT Time denotes the denoising inference time.}
    \begin{tabular}{lcccc}
    \toprule
    Method & \#Ref. Modality & \#Params & MLLM Time & DiT Time \\
    \midrule
    VACE~\cite{jiang2025vace}           & Text-Only     & 14B   & —             & $\sim$35 min \\
    Lucy-Edit~\cite{team2025lucy}       & Text-Only     & 5B    & —             & $\sim$4 min  \\
    DITTO~\cite{bai2025ditto}           & Text-Only     & 14B   & —             & $\sim$35 min \\
    NovaEdit~\cite{pan2026nova}         & Text-Only     & 1.3B  & —             & $\sim$15 min \\
    VideoCoF~\cite{yang2026videocof}    & Text-Only     & 14B   & —             & $\sim$35 min      \\
    Kiwi-Edit~\cite{lin2026kiwi}        & Text-Only     & 5B    &  $\sim$0.4s          & $\sim$4 min  \\
    \midrule
    \rowcolor{cyan!15} \textbf{VicEdit (Ours)} & Single Image & 5B & $\sim$0.5s & $\sim$4 min \\
    \rowcolor{cyan!15} \textbf{VicEdit (Ours)} & Image Pair   & 5B & $\sim$0.7s & $\sim$4 min \\
    \rowcolor{cyan!15} \textbf{VicEdit (Ours)} & Video Pair   & 5B & $\sim$1.0s & $\sim$4 min \\
    \bottomrule
    \end{tabular}
    \label{tab:computational_efficiency}
\end{table}

\begin{figure}[!t]
    \centering
    \includegraphics[width=0.8\linewidth]{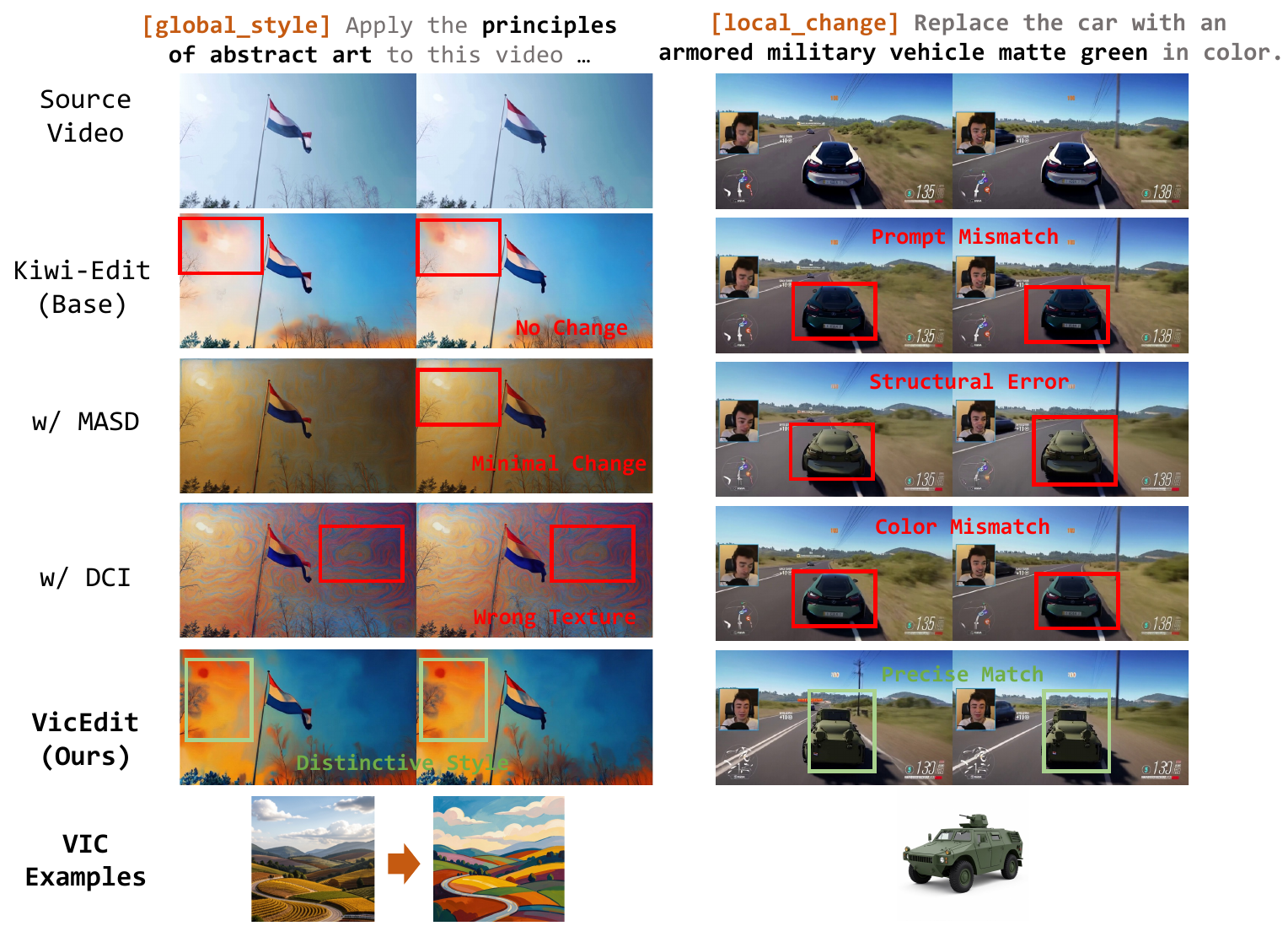}
    \caption{Qualitative ablation of MASD and DCI on (left) a global style editing task and (right) a local object replacement task. The full model better follows the reference style and editing instruction in the shown examples; the comparisons do not imply separately enforced roles for the two modules.}
    \label{fig:abla_visual}
\end{figure}

\subsection{Additional Ablation Results}
\label{suppl:visual_ablation}

Figure~\ref{fig:abla_visual} compares component variants on global style transfer and local object replacement. The variants differ in texture, color, and object appearance, while the full model better matches the demonstrated edits in these examples. These qualitative observations complement the numerical ablation; they do not establish an exclusive assignment of texture, structure, or instruction-following roles to individual modules.

\subsection{More Visualizations}

\begin{figure}[ht]
    \centering
    \includegraphics[width=\linewidth]{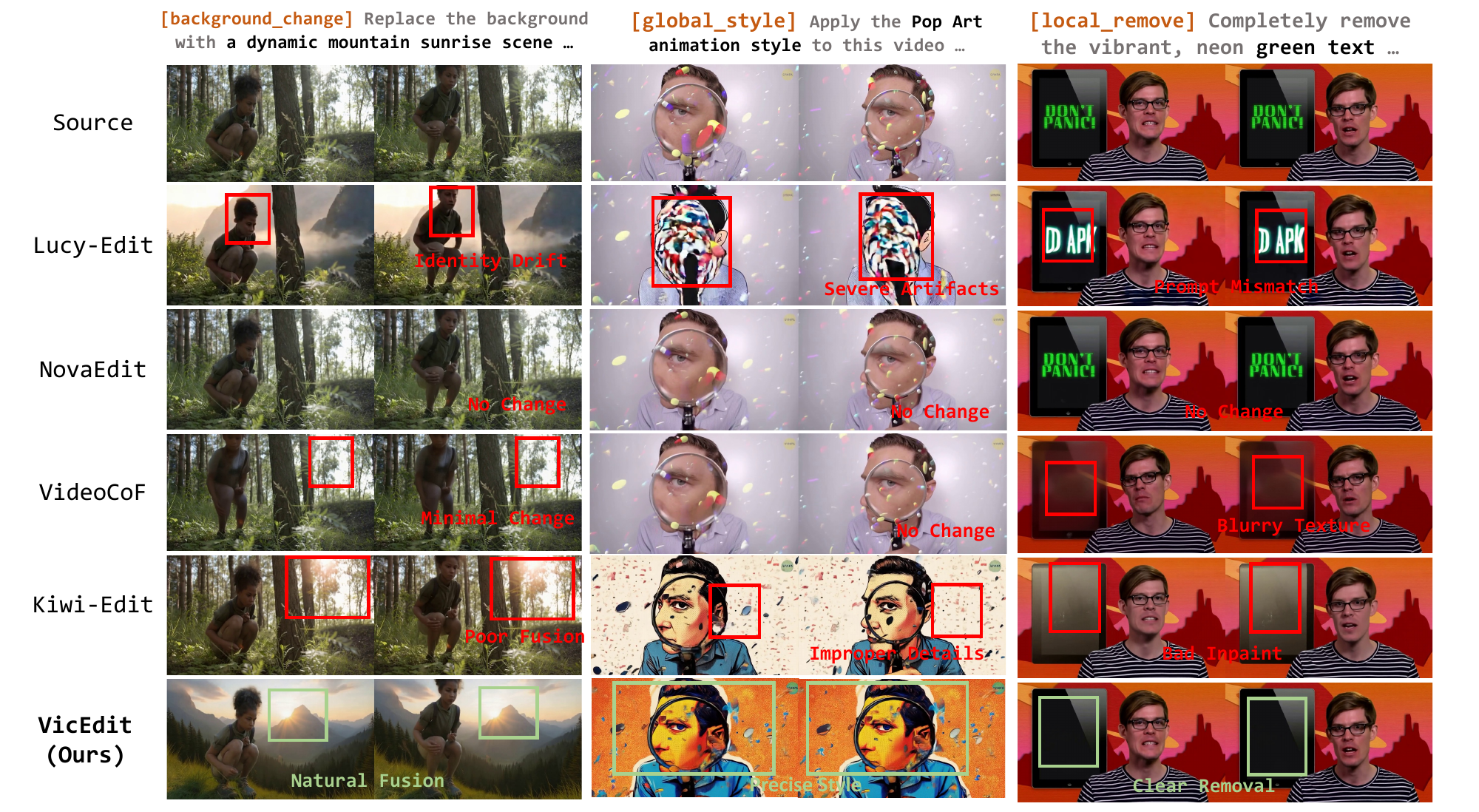}
    \caption{More comparisons with baselines on base editing tasks.}
    \label{fig:main_base_suppl}
\end{figure}

\begin{figure}[ht]
    \centering
    \includegraphics[width=\linewidth]{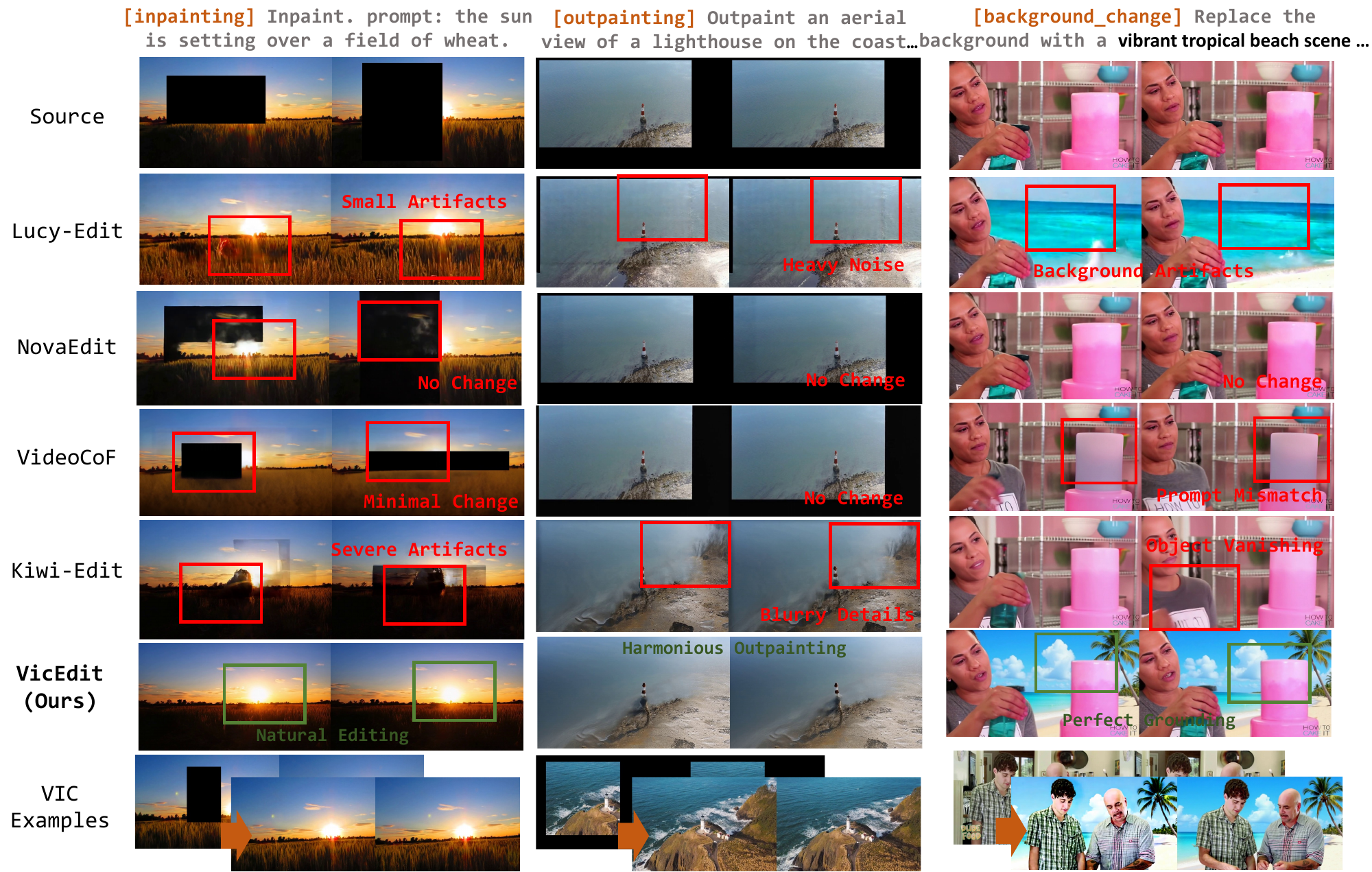}
    \caption{More comparisons with baselines on visual in-context editing tasks.}
    \label{fig:main_icl_suppl}
\end{figure}

We provide additional visual comparisons with baselines on both base editing and visual in-context editing tasks. As shown in Fig. \ref{fig:main_base_suppl} and Fig. \ref{fig:main_icl_suppl}, our method consistently outperforms existing approaches across a wider range of examples, further validating the effectiveness of VicEdit.

\newpage
\section*{NeurIPS Paper Checklist}

\begin{enumerate}

\item {\bf Claims}
    \item[] Question: Do the main claims made in the abstract and introduction accurately reflect the paper's contributions and scope?
    \item[] Answer: \answerYes{} % Replace by \answerYes{}, \answerNo{}, or \answerNA{}.
    \item[] Justification: The abstract and introduction clearly outline our primary contributions, including the proposed framework and the new dataset.

\item {\bf Limitations}
    \item[] Question: Does the paper discuss the limitations of the work performed by the authors?
    \item[] Answer: \answerYes{} % Replace by \answerYes{}, \answerNo{}, or \answerNA{}.
    \item[] Justification: We discuss the limitations and future work in the Conclusion section.

\item {\bf Theory assumptions and proofs}
    \item[] Question: For each theoretical result, does the paper provide the full set of assumptions and a complete (and correct) proof?
    \item[] Answer: \answerNA{} % Replace by \answerYes{}, \answerNo{}, or \answerNA{}.
    \item[] Justification: The paper does not involve intensive theoretical analysis.

    \item {\bf Experimental result reproducibility}
    \item[] Question: Does the paper fully disclose all the information needed to reproduce the main experimental results of the paper to the extent that it affects the main claims and/or conclusions of the paper (regardless of whether the code and data are provided or not)?
    \item[] Answer: \answerYes{}{} % Replace by \answerYes{}, \answerNo{}, or \answerNA{}.
    \item[] Justification: We have detailed the dataset specifications and training procedures in both the main paper and the appendix.

\item {\bf Open access to data and code}
    \item[] Question: Does the paper provide open access to the data and code, with sufficient instructions to faithfully reproduce the main experimental results, as described in supplemental material?
    \item[] Answer: \answerNo{} % Replace by \answerYes{}, \answerNo{}, or \answerNA{}.
    \item[] Justification: Cleaning the large-scale dataset and codebase for public release is underway. We commit to making all code and data publicly available upon acceptance to ensure reproducibility.

\item {\bf Experimental setting/details}
    \item[] Question: Does the paper specify all the training and test details (e.g., data splits, hyperparameters, how they were chosen, type of optimizer) necessary to understand the results?
    \item[] Answer: \answerYes{} % Replace by \answerYes{}, \answerNo{}, or \answerNA{}.
    \item[] Justification: We have detailed the dataset specifications and training procedures in both the main paper and the appendix.

\item {\bf Experiment statistical significance}
    \item[] Question: Does the paper report error bars suitably and correctly defined or other appropriate information about the statistical significance of the experiments?
    \item[] Answer: \answerYes{} % Replace by \answerYes{}, \answerNo{}, or \answerNA{}.
    \item[] Justification: We report the results of three runs for each experiment.

\item {\bf Experiments compute resources}
    \item[] Question: For each experiment, does the paper provide sufficient information on the computer resources (type of compute workers, memory, time of execution) needed to reproduce the experiments?
    \item[] Answer: \answerYes{} % Replace by \answerYes{}, \answerNo{}, or \answerNA{}.
    \item[] Justification: We provide the total computation hours and the number of devices used for training in the Appendix, ensuring reproducibility without disclosing specific infrastructure details.
    
\item {\bf Code of ethics}
    \item[] Question: Does the research conducted in the paper conform, in every respect, with the NeurIPS Code of Ethics \url{https://neurips.cc/public/EthicsGuidelines}?
    \item[] Answer: \answerYes{} % Replace by \answerYes{}, \answerNo{}, or \answerNA{}.
    \item[] Justification: Our research follows the NeurIPS Code of Ethics.

\item {\bf Broader impacts}
    \item[] Question: Does the paper discuss both potential positive societal impacts and negative societal impacts of the work performed?
    \item[] Answer: \answerYes{} % Replace by \answerYes{}, \answerNo{}, or \answerNA{}.
    \item[] Justification: We discuss the potential positive impacts on video creation and the negative risks.
    
\item {\bf Safeguards}
    \item[] Question: Does the paper describe safeguards that have been put in place for responsible release of data or models that have a high risk for misuse (e.g., pre-trained language models, image generators, or scraped datasets)?
    \item[] Answer: \answerYes{} % Replace by \answerYes{}, \answerNo{}, or \answerNA{}.
    \item[] Justification: We have implemented semantic filtering during data curation to remove unsafe content and will include usage restrictions in our license.

\item {\bf Licenses for existing assets}
    \item[] Question: Are the creators or original owners of assets (e.g., code, data, models), used in the paper, properly credited and are the license and terms of use explicitly mentioned and properly respected?
    \item[] Answer: \answerYes{} % Replace by \answerYes{}, \answerNo{}, or \answerNA{}.
    \item[] Justification: We properly credit and cite Senorita, OpenVE, and Wan-TI2V. All assets are used according to their respective licenses.

\item {\bf New assets}
    \item[] Question: Are new assets introduced in the paper well documented and is the documentation provided alongside the assets?
    \item[] Answer: \answerYes{} % Replace by \answerYes{}, \answerNo{}, or \answerNA{}.
    \item[] Justification: We introduce \dataset{} and \benchmark{}, providing documentation and maintenance plans in the Appendix.

\item {\bf Crowdsourcing and research with human subjects}
    \item[] Question: For crowdsourcing experiments and research with human subjects, does the paper include the full text of instructions given to participants and screenshots, if applicable, as well as details about compensation (if any)? 
    \item[] Answer: \answerNo{} % Replace by \answerYes{}, \answerNo{}, or \answerNA{}.
    \item[] Justification: We conducted a human preference study with ten annotators. Appendix~\ref{suppl:evaluation} describes the evaluation protocol and scoring procedure. 

\item {\bf Institutional review board (IRB) approvals or equivalent for research with human subjects}
    \item[] Question: Does the paper describe potential risks incurred by study participants, whether such risks were disclosed to the subjects, and whether Institutional Review Board (IRB) approvals (or an equivalent approval/review based on the requirements of your country or institution) were obtained?
    \item[] Answer: \answerNo{} % Replace by \answerYes{}, \answerNo{}, or \answerNA{}.
    \item[] Justification: The paper includes a human preference study, but does not report participant risk disclosures or the status of institutional ethics review or exemption.

\item {\bf Declaration of LLM usage}
    \item[] Question: Does the paper describe the usage of LLMs if it is an important, original, or non-standard component of the core methods in this research? Note that if the LLM is used only for writing, editing, or formatting purposes and does \emph{not} impact the core methodology, scientific rigor, or originality of the research, declaration is not required.
    %this research? 
    \item[] Answer: \answerYes{} % Replace by \answerYes{}, \answerNo{}, or \answerNA{}.
    \item[] Justification: MLLMs (\textit{e.g.}, Qwen-VL) are used as a core component in our data construction pipeline for automated instruction generation and verification.

\end{enumerate}

\end{document}